\documentclass[journal,10pt]{IEEEtran}
\pdfoutput=1

\newlength{\figwidth}
\usepackage{color}
\usepackage[nosort]{cite}        
\usepackage{url} 
\usepackage[intlimits]{amsmath}
\usepackage{bbm}
\usepackage{graphicx}
\usepackage{paralist}
\usepackage{fancyref}
\usepackage[stretch=16,shrink=16,step=4]{microtype}
\usepackage{vmr-symbols-vecbold}
\usepackage{standard-macros}
\makeatletter
\def\@IEEEinterspaceratioM{0.265}
\def\@IEEEinterspaceMINratioM{0.1651}
\def\@IEEEinterspaceMAXratioM{0.38}

\def\@IEEEinterspaceratioB{0.31}
\def\@IEEEinterspaceMINratioB{0.19}
\def\@IEEEinterspaceMAXratioB{0.38}
\@IEEEtunefonts
\makeatother
\newcommand{\trainingdata}{\boldsymbol{Z}}

\newcommand{\supersample}{\tilde{\boldsymbol{Z}}}

\newcommand{\subsetchoice}{\boldsymbol{S}}

\newcommand{\genop}{\textnormal{gen}}
\newcommand{\gen}{\genop(W,\trainingdata)}

\newcommand{\genwZ}{\genop(w,\trainingdata)}

\newcommand{\genS}{\genop(W,\trainingdata(\subsetchoice))}
\newcommand{\genShat}{\widehat\genop(W,\supersample, \subsetchoice)}
\newcommand{\genShatwz}{\widehat\genop(w,\widetilde{\boldsymbol{z}}, \subsetchoice)}

\newcommand{\jointdistro}{P_{W\! \supersample\! \subsetchoice }}

\newcommand{\infdensop}{\imath}
\newcommand{\infdens}{\infdensop(W,\trainingdata)}
\newcommand{\condinfdens}{\infdensop(W,\subsetchoice\vert \supersample)}

\newtheorem{remark}{Remark}

\begin{document}

\IEEEoverridecommandlockouts

\title{Generalization Bounds via Information Density and Conditional Information Density}
\author{Fredrik Hellstr\"om,~\IEEEmembership{Student Member,~IEEE}, Giuseppe~Durisi,~\IEEEmembership{Senior Member,~IEEE}
\thanks{This work was partly supported by the Wallenberg AI, Autonomous Systems and Software Program (WASP) funded by the Knut and Alice Wallenberg Foundation.}
\thanks{Parts of the material of this paper were presented at the International Symposium on Information Theory (ISIT), June 2020, Los Angeles, CA~\cite{PA-hellstrom-20a}}.
\thanks{F. Hellstr\"om and G. Durisi are with the Department of Electrical Engineering, Chalmers University of Technology, Gothenburg, Sweden, (e-mail: \{frehells,durisi\}@chalmers.se).
}
}%

\maketitle
\begin{abstract}
We present a general approach, based on exponential inequalities, to derive bounds on the generalization error of randomized learning algorithms. Using this approach, we provide bounds on the average generalization error as well as bounds on its tail probability, for both the PAC-Bayesian and single-draw scenarios. Specifically, for the case of sub-Gaussian loss functions, we obtain novel bounds that depend on the information density between the training data and the output hypothesis. When suitably weakened, these bounds recover many of the information-theoretic bounds available in the literature. We also extend the proposed exponential-inequality approach to the setting recently introduced by Steinke and Zakynthinou~(2020), where the learning algorithm depends on a randomly selected subset of the available training data. For this setup, we present bounds for bounded loss functions in terms of the conditional information density between the output hypothesis and the random variable determining the subset choice, given all training data. Through our approach, we recover the average generalization bound presented by Steinke and Zakynthinou~(2020) and extend it to the PAC-Bayesian and single-draw scenarios. For the single-draw scenario, we also obtain novel bounds in terms of the conditional $\alpha$-mutual information and the conditional maximal leakage.
\end{abstract}

\section{Introduction}\label{sec-a:introduction}
A randomized learning algorithm $P_{W\vert \trainingdata}$ consists of a probabilistic mapping from a set of training data $\trainingdata=(Z_1,\dots,Z_n)\in \mathcal{Z}^n$, which we assume to have been generated independently from an unknown distribution $P_Z$ on the instance space $\mathcal{Z}$, to an output hypothesis $W\in\mathcal{W}$, where $\mathcal{W}$ is the hypothesis space. The goal is to find a hypothesis $W$ that results in a small expected loss $L_{P_Z}(W)= \Exop_{P_Z}[\ell(W,Z)]$, where $\ell(\cdot,\cdot)$ is some suitably chosen loss function. A typical strategy to achieve this goal is \emph{empirical risk minimization}, according to which $W$ is selected so as to minimize the empirical loss $L_{\trainingdata}(W)=\tfrac{1}{n}\sum_{i=1}^n\ell(W,Z_i)$. A central objective in statistical learning theory is to determine when this choice results in a small population loss $L_{P_Z}(W)$. To this end, one seeks to bound the \emph{generalization error}, defined as $\textnormal{gen}(W,\trainingdata) = L_{P_Z}(W)-L_{\trainingdata}(W)$. Since the learning algorithm is randomized, bounds on $\gen$ can come in several flavors. One possibility is to bound the average generalization error $\abs{\Exop_{P_{W\!\trainingdata}}[\gen]}$. In practice, one might be more interested in an upper bound on $\abs{\Exop_{P_{W\vert \trainingdata}}[\gen]}$ that holds with probability at least $1-\delta$ under the product distribution $P_{\trainingdata}$. Here, $\delta\in(0,1)$ is the so-called confidence parameter. Bounds of this type, which are typically referred to as \emph{probably approximately correct (PAC)-Bayesian} bounds~\cite{PA-mcallester98-07a,PA-guedj19-01a}, are relevant for the scenario in which a new hypothesis $W$ is drawn from $P_{W\vert \trainingdata}$ every time the algorithm is used. For the scenario in which $W$ is drawn from $P_{W\vert \trainingdata}$ only once---a setup that, following the terminology in~\cite{PA-catoni07-a}, we shall refer to as \emph{single-draw}---one may instead be interested in obtaining an upper bound on $\abs{\gen}$ that holds with probability at least $1-\delta$ under the joint distribution $P_{W\!\trainingdata}$. If the dependence of a probabilistic bound (PAC-Bayesian or single-draw) on $\delta^{-1}$ is at most logarithmic, the bound is usually referred to as a \emph{high-probability} bound. Furthermore, a probabilistic bound is termed \textit{data-independent} if it does not depend on the specific instance of $\trainingdata$, and \textit{data-dependent} if it does. Data-independent bounds allow one to characterize the sample complexity~\cite[p.~44]{PA-shalev-shwartz14-a}, defined as the minimum number of training samples needed to guarantee that the generalization error is within a desired range, with a desired confidence level. However, data-dependent results are often tighter. Indeed, many of the available data-independent bounds can be recovered as relaxed versions of data-dependent bounds.

Classical PAC bounds on the generalization error, such as those based on the Vapnik-Chervonenkis (VC) dimension~\cite[p.~67]{PA-shalev-shwartz14-a}, are probabilistic bounds of a stronger variety than the PAC-Bayesian and single-draw bounds just introduced. Indeed, they hold uniformly for \emph{all} $w\in \mathcal{W}$ under $P_{\trainingdata}$. As a consequence, these bounds depend on structural properties of the hypothesis class $\mathcal{W}$ rather than on properties of the algorithm, and tend to be crude when applied to modern machine learning algorithms~\cite{PA-Zhang-16}.

\paragraph*{Prior Work}
By generalizing a result obtained in~\cite{PA-russo16-05b} in the context of adaptive data analysis, Xu and Raginsky~\cite{PA-xu17-05a} obtained a bound on the average generalization error in terms of the mutual information $I(W;\trainingdata)$ between the output hypothesis $W$ and the training data $\trainingdata$.
A drawback of the bound in~\cite{PA-xu17-05a} is that it is vacuous whenever the joint distribution $P_{W\!\trainingdata}$ is not absolutely continuous with respect to $P_W\!P_{\trainingdata}$, the product of the marginal distributions of $W$ and $\trainingdata$.
This occurs, for example, when $W$ is given by a deterministic function of $\trainingdata$, and $W$ and $\trainingdata$ are separately continuous random variables. In~\cite{PA-Bu-19-ISIT}, Bu \textit{et al.} rectified this by obtaining a tighter bound in terms of the individual-sample mutual information $I(W;Z_i)$, which can be bounded even when $I(W;\trainingdata)=\infty$. In~\cite{PA-Asadi2018}, Asadi \textit{et al.} combined the mutual information bound with the chaining technique~\cite{PA-vanHandel-16}, which exploits structure in the hypothesis class to tighten bounds. In some cases, this is shown to give stronger bounds than either the mutual information bound or the chaining bound individually.

To be evaluated, all of the aforementioned bounds require knowledge of the marginal distribution $P_W$, which depends on the data distribution $P_Z$. In practice, this data distribution is unknown, making the marginal $P_W$ intractable. In light of this, Achille and Soatto~\cite{PA-Achille-18} provided an upper bound on the mutual information between the training data and the output hypothesis in terms of the relative entropy between $P_{W\vert\trainingdata}$ and a fixed, auxiliary distribution on the hypothesis space $\mathcal{W}$, and showed that this results in a computable upper bound on the average generalization error. Similarly, Negrea \textit{et al.}~\cite{PA-Negrea2019} provided generalization bounds in terms of an auxiliary, possibly data-dependent distribution on $\mathcal{W}$. This weakens the bound, but makes it computable. Their use of the expected square root of the relative entropy $\relent{P_{W\vert \trainingdata}}{P_W}$, which they call \textit{disintegrated} mutual information, in place of the mutual information leads to further improvements on the basic bound.

Recent studies, starting with the work of Steinke and Zakynthinou~\cite{PA-steinke20-a}, have considered a setting with more structure, where it is assumed that a set $\supersample$ consisting of $2n$ independent and identically distributed (\iid) training samples from $P_Z$ is available, and that $\trainingdata$ is formed by selecting $n$ entries of $\supersample$ at random.
Since the motivation for this structure is to derive generalization bounds in terms of a \textit{conditional} mutual information (CMI), we refer to this setup as the \textit{CMI setting}, and call the setting without this additional structure the \textit{standard setting}.
In the CMI setting, as aforementioned, the average generalization error can be bounded by a quantity that depends on the CMI between the output hypothesis and the random variable that determines the selected training data $\trainingdata$, given $\supersample$ \cite[Thm.~5.1]{PA-steinke20-a}.
One advantage of this bound over the standard mutual information bound is that the CMI is always bounded. This broadens the applicability of the bound and results in tighter estimates.
Also, as discussed in~\cite[Sec.~4]{PA-steinke20-a}, the CMI that appears in the bound has strong connections to classical generalization measures, such as VC dimension, compressibility, and stability.
In~\cite{PA-Haghifam2020}, Haghifam \textit{et al.} provided an individual-sample strengthening of this result, as well as improvements through their use of disintegration.
In all of these derivations, the loss function is required to be bounded, which is a stronger requirement than what is needed in the standard setting.

All of the information-theoretic results discussed so far pertain to bounds on the average generalization error. In~\cite[App.~A.3]{PA-bassily18-02a}, Bassily \textit{et al.} provided a PAC-Bayesian bound in terms of mutual information. This bound is essentially a data-independent relaxation of a well-known data-dependent bound from the PAC-Bayesian literature~\cite{PA-guedj19-10a}. The dependence of the original data-dependent PAC-Bayesian bound on the confidence parameter $\delta$ is of order $\log(1/\delta)$, making it a high-probability bound. However, its mutual information relaxation in~\cite{PA-bassily18-02a} has a less benign $1/\delta$-dependence. PAC-Bayesian techniques have recently found some success in producing non-vacuous generalization bounds for (randomized) deep neural networks. In~\cite{PA-Dziugaite2017}, Dziugaite and Roy optimized a PAC-Bayesian bound to get non-vacuous generalization estimates for a simple neural network setup. 
These estimates were recently further improved in~\cite{PA-dziugaite-20}.
In~\cite{PA-zhou2018nonvacuous}, Zhou \textit{et al.} derived a bound for compressed networks, i.e., small neural networks that are formed by pruning larger ones, and illustrated numerically that the bound is non-trivial for realistic settings. An extensive survey of the vast PAC-Bayesian literature, which is beyond the scope of this paper, can be found in, e.g.,~\cite{PA-guedj19-01a}.

Finally, we survey the single-draw bounds that are relevant for our discussion. In addition to the aforementioned average and PAC-Bayesian bounds, both Xu and Raginsky~\cite[Thm.~3]{PA-xu17-05a} and Bassily \textit{et al.}~\cite{PA-bassily18-02a} also provided single-draw generalization bounds in terms of mutual information. For both of them, the dependence on $\delta$ is of order $1/\delta$. In~\cite{PA-esposito19-12a}, Esposito \textit{et al.} provided bounds in terms of a whole host of information-theoretic quantities, such as the R\'enyi divergence, the $\alpha$-mutual information, and the maximal leakage. An interesting aspect of their $\alpha$-mutual information bound is that, unlike the mutual information bounds in~\cite[Thm.~3]{PA-xu17-05a} and~\cite{PA-bassily18-02a}, it is a high-probability bound. However, this bound does not imply a stronger mutual information bound. Indeed, if one lets $\alpha \rightarrow 1$, for which the $\alpha$-mutual information reduces to the ordinary mutual information, the bound becomes vacuous. Bounds on the average generalization error are also provided~\cite[Sec.~III.D]{PA-esposito19-12a}, but these are generally weaker than the mutual information bounds in~\cite{PA-xu17-05a}. In the same vein, Dwork \textit{et al.} derived single-draw generalization bounds in terms of other algorithmic stability measures, such as differential privacy~\cite{PA-Dwork-14-wa} and (approximate) max-information~\cite{PA-dwork15-06a}. These bounds are of the high-probability variety, but are typically weaker than the aforementioned maximal leakage bound~\cite[Sec.~V]{PA-esposito19-12a}. All of the single-draw bounds mentioned here are data-independent.

\paragraph*{Contributions}
In this paper, we derive bounds of all three flavors---average, PAC-Bayesian, and single-draw---for both the standard setting and the CMI setting. In the standard setting, we use the sub-Gaussianity of the loss function, together with a change of measure argument, to obtain exponential inequalities in terms of the \emph{information density} between the hypothesis $W$ and the training data $\trainingdata$. 
These exponential inequalities provide a framework that can be used not only to derive novel bounds, but also to recover several known results, which were originally derived using a host of different tools.
In this sense, it provides a unifying approach for deriving information-theoretic generalization bounds.
Through simple manipulations of the exponential inequalities, we recover the average generalization bound in~\cite[Thm.~1]{PA-xu17-05a} and the data-dependent PAC-Bayesian bound in~\cite[Prop.~3]{PA-guedj19-10a}.
We also derive a novel data-dependent single-draw bound.
Moreover, we obtain two novel data-independent bounds that are explicit in the $t$th moments of the relative entropy~$\relent{P_{W\vert \trainingdata}}{P_W}$ and of the information density, respectively.
The dependence of these bounds on the confidence parameter $\delta$ is of order $1/\delta^t$. This is more favorable than that of similar bounds reported in~\cite{PA-xu17-05a} and~\cite{PA-bassily18-02a}, which have a dependence of order $1/\delta$.
The moment bounds that we obtain illustrate that tighter estimates of the generalization error are available with higher confidence if the higher moments of the information measures that the bounds depend on are sufficiently small.
Through a more refined analysis, we also obtain a high-probability data-independent single-draw bound in terms of maximal leakage.
This result coincides with~\cite[Cor.~10]{PA-esposito19-12a}, up to a logarithmic term. Finally, by using a different approach that relies on tools from binary hypothesis testing, we obtain a data-independent single-draw bound in terms of the tail of the information density.
Similarly to the moment bounds, this bound illustrates that the faster the decay of the tail of the information density random variable, the more benign the dependence of the bound on $\delta$.

Moving to the CMI setting, we establish exponential inequalities, similar to those for the standard setting, in terms of the~\emph{conditional information density} between the hypothesis and a random variable that selects the data to be used for training, given all data samples. These exponential inequalities are derived under the more stringent assumption of a bounded loss function.
Then, we use these inequalities to reobtain the average generalization bound in~\cite[Cor.~5.2]{PA-steinke20-a}, and to derive novel PAC-Bayesian and single-draw bounds, both of data-dependent and of data-independent flavor.
Similarly to the standard setting, we also obtain a bound that is explicit in the tail of the conditional information density by using tools from binary hypothesis testing.
Finally, inspired by~\cite{PA-esposito19-12a}, we derive a parametric inequality that can be used to obtain data-independent single-draw bounds. Using this inequality, we extend the results in~\cite{PA-esposito19-12a} for bounded loss functions to the CMI setting, and obtain bounds in terms of the conditional versions of the $\alpha$-mutual information, the R\'enyi divergence, and the maximal leakage. Under some conditions, the conditional maximal leakage bound turns out to be stronger than its maximal leakage counterpart.

\section{Preliminaries}\label{sec-a:preliminaries}
In this section, we introduce some notation, define relevant information-theoretic quantities, and present some general results that will be used repeatedly in the remainder of this paper.

\paragraph*{Standard and CMI settings} Let $\mathcal{Z}$ be the instance space, $\mathcal{W}$ be the hypothesis space, and $\ell:\mathcal{W}\times\mathcal{Z}\rightarrow \reals^+$ be the loss function. In the standard setting, $n$ training samples $\trainingdata=(Z_1,\dots,Z_n)$ are available. These $n$ samples constitute the training data. We assume that all entries of $\trainingdata$ are drawn independently from some unknown distribution $P_Z$ on $\mathcal{Z}$. In the CMI setting, $2n$ training samples $\supersample=(\tilde Z_1,\dots,\tilde Z_{2n})$ are available, with all entries of $\supersample$ being drawn independently from $P_Z$. However, only a randomly selected subset of cardinality $n$ is actually used as the training data. Following~\cite{PA-steinke20-a}, we assume that the training data $\trainingdata(\subsetchoice)$ is selected as follows. Let $\subsetchoice=(S_1,\dots,S_n)$ be an $n$-dimensional random vector, the elements of which are drawn independently from a $\mathrm{Bern}(1/2)$ distribution and are independent of $\supersample$. Then, for $i=1,\dots,n$, the $i$th training sample in $\trainingdata(\subsetchoice)$ is $Z_i(S_i)=\tilde Z_{i+S_in}$. A randomized learning algorithm is a conditional distribution $P_{W\vert \trainingdata}$. We let $L_{\trainingdata}(W)=\tfrac{1}{n}\sum_{i=1}^n\ell(W,Z_i)$ denote the empirical loss and $L_{P_Z}(W)=\Exop_{P_Z}[\ell(W,Z)]$ the population loss. The generalization error is defined as $\gen = L_{P_Z}(W)-L_{\trainingdata}(W)$. 

\paragraph*{Information Measures}%
A quantity that will appear in many of our bounds is the information density, defined as
\begin{equation}\label{B_eq:def_info_dens}
\imath(W,\trainingdata) = \log \frac{\dv P_{W\!\trainingdata} }{\dv P_W\!P_{\trainingdata}}
\end{equation}
where $\dv P_{W\!\trainingdata}/\dv P_W\!P_{\trainingdata}$ is the Radon-Nikodym derivative of $P_{W\!\trainingdata}$ with respect to $P_W\!P_{\trainingdata}$. Here, $P_W$ is the distribution induced on the hypothesis space $\mathcal W$ by $P_Z$ through $P_{W\vert \trainingdata}$. The information density is well defined whenever $P_{W\!\trainingdata}$ is absolutely continuous with respect to $P_W\!P_{\trainingdata}$, which we denote by $P_{W\!\trainingdata}\ll P_W\!P_{\trainingdata}$. The name information density is motivated by the fact that its expectation under $P_{W\!\trainingdata}$ is the mutual information $I(W;\trainingdata)$. In the CMI setting, several of our bounds will be in terms of the conditional information density
\begin{equation}\label{B_eq:conditional_info_density}
\imath(W,\subsetchoice\vert \supersample) = \log \frac{\dv P_{W\! \supersample\! \subsetchoice} }{\dv P_{W\vert \supersample  }P_{\supersample\!\subsetchoice} }
\end{equation}
where $P_{W\vert \supersample}$ is a conditional distribution on $\mathcal{W}$ given $\supersample$, obtained by marginalizing out $\subsetchoice$. Here, the absolute continuity requirement is that $P_{W\!\supersample\!\subsetchoice}\ll P_{W\vert\supersample}P_{\supersample\!\subsetchoice}$. In the CMI setting, this is satisfied since $P_{W\vert \supersample}$ is obtained by marginalizing out the discrete random variable $\subsetchoice$ from $P_{W\vert \supersample\!\subsetchoice}$. If we take the expectation of $\imath(W,\subsetchoice\vert \supersample)$ under the joint distribution $P_{W\!\supersample\!\subsetchoice}$, we obtain the CMI $I(W;\subsetchoice\vert \supersample)$, a key quantity in the bounds developed in~\cite{PA-steinke20-a}.

Let $\alpha\in (0,1)\cup(1,\infty)$. The R\'enyi divergence of order $\alpha$ is defined as~\cite{PA-VanErven2014}
\begin{equation}\label{B_eq:def_renyi_divergence}
\alpharelent{\alpha}{P_{W\!\trainingdata}}{P_W\!P_{\trainingdata}}\!=\!\frac{1}{\alpha\!-\!1} \!\log \Exop_{P_W\!P_{\trainingdata}}\lefto[\exp\lefto(\alpha\infdens\right)\right].\!
\end{equation}
In the limit $\alpha\rightarrow 1$, it reduces to the relative entropy $\relent{P_{W\!\trainingdata}}{P_W\!P_{\trainingdata}}$. The conditional R\'enyi divergence of order $\alpha$ is given by~\cite{PA-verdu15-02a}
\begin{multline}\label{B_eq:def_cond_renyi_divergence}
\alphaconrelent{\alpha}{P_{W\vert \supersample\!\subsetchoice}P_{\subsetchoice\vert \supersample} }{P_{W\vert \supersample}P_{\subsetchoice\vert \supersample }}{P_{\supersample}} \\
= \frac{1}{\alpha-1}\log \Exop_{P_ {\supersample}P_{W\vert \supersample }P_{\subsetchoice\vert \supersample}}\lefto[\exp\lefto(\alpha\condinfdens \right) \right].
\end{multline}
The $\alpha$-mutual information, which is studied in depth in \cite{PA-verdu15-02a}, is defined as%
\begin{equation}\label{B_eq:def_alpha_MI}
I_\alpha(\trainingdata;W) = \frac{1}{\alpha-1}\log \Exop_{P_W}^\alpha\lefto[ \Exop_{P_{\trainingdata} }^{1/\alpha} \lefto[\exp\lefto(\alpha\infdens\right)  \right] \right].
\end{equation}
In the limit $\alpha\rightarrow 1$, it reduces to the mutual information $I(W;\trainingdata)$, whereas for $\alpha\rightarrow \infty$, it becomes the maximal leakage \cite{PA-issa16-a}:
\begin{equation}\label{B_eq:def_maximal_leakage}
\mathcal{L}(\trainingdata\rightarrow W)=\log \Exop_{P_W}\lefto[ \esssup_{P_{\trainingdata}}\exp\lefto(\infdens\right)  \right].
\end{equation}
Here, the essential supremum of a measurable function $f(\cdot)$ of a random variable $\trainingdata$ distributed as $P_{\trainingdata}$ is defined as
\begin{equation}
\esssup_{P_{\trainingdata}} f(\trainingdata) = \inf_{a\in \reals}\biggo[P_{\trainingdata}\big(\{\trainingdata: f(\trainingdata)>a\}\big)=0 \bigg] .
\end{equation}

The conditional $\alpha$-mutual information does not have a commonly accepted definition. In~\cite{PA-Tomamichel2018}, three definitions are provided and given operational interpretations, two of which have known closed-form expressions. The first coincides with the conditional R\'enyi divergence, while the second, which we will term $I_\alpha(W;\subsetchoice\vert \supersample)$, is defined as
\begin{multline}\label{B_eq:def_cond_alpha_MI}
I_\alpha(W; \boldsymbol{S}\vert  \tilde{\boldsymbol{Z}} )\\
=\frac{1}{\alpha\!-\!1}\log \Exop_{P_ {\tilde{\boldsymbol{Z}}}}\!\lefto[\Exop_{P_{W\vert \tilde{\boldsymbol{Z}} } }^{\alpha}\!\lefto[ \Exop_{P_{ \boldsymbol{S}\vert \tilde{\boldsymbol{Z}}}}^{1/\alpha}
\lefto[\exp\lefto(\alpha\condinfdens \!\right) \right]\right]\right]\!.\!
\end{multline}
In the limit $\alpha\rightarrow \infty$, this reduces to the conditional maximal leakage\cite[Thm.~6]{PA-issa16-a}
\begin{equation}\label{B_eq:def_cond_maximal_leakage}
 \mathcal{L}(\subsetchoice\! \rightarrow\! W\vert \supersample)\!=\!\log\esssup_{P_{\supersample}} \Exop_{P_{W\vert \supersample}}\!\lefto[\! \esssup_{P_{\subsetchoice\vert \supersample} }e^{\condinfdens} \right].\!
\end{equation}
Note that $\subsetchoice$ and $\supersample$ are independent in the CMI setting. Hence, $P_{\subsetchoice\vert \supersample}$ can be replaced by $P_{\subsetchoice}$ in~\eqref{B_eq:def_cond_renyi_divergence},~\eqref{B_eq:def_cond_alpha_MI}, and~\eqref{B_eq:def_cond_maximal_leakage}.

\paragraph*{Useful Results}Many previous studies have used the data-processing inequality as a tool for deriving generalization bounds~\cite{PA-bassily18-02a,PA-esposito19-12a}. In binary hypothesis testing, it is known that the data-processing inequality only provides weak converse bounds on the region of achievable error rates. To get strong converse bounds, one relies on the following lemma instead~\cite[Lem.~12.2]{PA-polyanskiy19-a}.
\begin{lem}[Strong Converse Lemma]\label{lem:strong_converse_lemma}
Let $P$ and $Q$ be probability distributions on some common space $\mathcal{X}$ such that $P$ is absolutely continuous with respect to $Q$, and let $\setE\in \mathcal{X}$ be a measurable set.
Then, for all $\gamma\in\reals$,
\begin{equation}\label{B_eq:lem_strong_converse_lemma}
    P[\setE] \leq P\lefto[\log \frac{\dv P}{\dv Q} >  \gamma \right] + e^\gamma Q[\setE].
\end{equation}
\end{lem}
In Section~\ref{sec-a:uncond_sd_strong_conv} and Section~\ref{sec-a:cond_sd_strong_conv}, we will show how to use this result to derive generalization bounds.

We will also make repeated use of the following concentration results for sub-Gaussian random variables.
\begin{lem}[Sub-Gaussian concentration]\label{lem:hoeffdings_inequality}
Let $X\distas P_X$ be a $\sigma$-sub-Gaussian random variable, i.e., a random variable satisfying the following inequality for all $\lambda\in\reals$:~\cite[Prop.~2.5]{PA-wainwright19-a}
\begin{equation}\label{B_eq:lem_hoeffdings_inequality}
\Exop\lefto[\exp(\lambda(X-\Exop[X]) ) \right] \leq \exp\lefto(\frac{\lambda^2\sigma^2}{2}\right).
\end{equation}
Then, for all $\epsilon>0$,
\begin{equation}
P_X\lefto(\abs{X-\Exop[X]}\geq \epsilon \right) \leq 2\exp\lefto(-\frac{\epsilon^2}{2\sigma^2} \right).
\end{equation}
Furthermore, for all~$\lambda\in[0,1)$~\cite[Thm.~2.6.(IV)]{PA-wainwright19-a},
\begin{equation}\label{B_eq:subgauss-square-lemma}
\Ex{}{\exp\lefto(\frac{\lambda X^2}{2\sigma^2} \right) } \leq \frac{1}{\sqrt{1-\lambda}}.
\end{equation}
Note that a random variable bounded on $[a,b]$ is $\sigma$-sub-Gaussian with $\sigma=(b-a)/2$. Also, if $X_i$, for $i=1,\dots,n$, are independent $\sigma$-sub-Gaussian random variables, the average $(1/n)\sum_{i=1}^n X_i$ is $\sigma/\sqrt{n}$-sub-Gaussian.
\end{lem}
\section{Generalization Bounds for the Standard Setting}\label{sec-a:uncond_results}
In this section, we study the standard setting described in Section~\ref{sec-a:preliminaries}. We will assume that the loss function $\ell(w,Z)$ is $\sigma$-sub-Gaussian under $P_Z$ for all $w\in\mathcal{W}$. This means that, for all $\lambda\in\reals$ and for all $w\in\mathcal{W}$,
\begin{equation}\label{B_eq:sub-Gaussian_def}
\Exop_{P_Z}\lefto[ \exp\lefto(\lambda(\Exop_{P_Z}\lefto[\ell(w,Z)\right]\!-\!\ell(w,\!Z) \right) \right]\!\leq\! \exp\lefto(\!\frac{\lambda^2\sigma^2}{2}\!\right)\!.\!
\end{equation}
We will derive bounds on the generalization error of a probabilistic learning algorithm $P_{W\vert \trainingdata}$ in terms of some function of the information density~\eqref{B_eq:def_info_dens}. As previously mentioned, several different notions of generalization error bounds have been investigated in the literature. One such notion is that of average generalization bounds, where we want to find an $\epsilon$ such that
\begin{equation}\label{B_eq:uncond_exp_generalization_bound_def}
\abs{\Exop_{P_{W\!\trainingdata}}\lefto[\gen \right]} \leq \epsilon.
\end{equation}
This $\epsilon$ will in general depend on the joint distribution $P_{W\!\trainingdata}$, on properties of the loss function, and on the cardinality $n$ of the training data. We will study this type of bounds in Section~\ref{sec-a:uncond_exp}.

Another approach, typically studied in the PAC-Bayesian literature, is to find probabilistic bounds of the following form: with probability at least $1-\delta$ under $P_{\trainingdata}$,
\begin{equation}\label{B_eq:uncond_pacb_generalization_bound_def}
\abs{\Exop_{P_{W\vert \trainingdata}}\lefto[\gen \right]} \leq \epsilon.
\end{equation}
This bound is interesting when we have a fixed data set $\trainingdata$, but draw a new hypothesis according to $P_{W\vert \trainingdata}$ each time we want to use our algorithm. 
The main advantage of this PAC-Bayesian approach is that, by considering a distribution over the hypothesis class rather than just a single hypothesis, one can capture uncertainty about the hypotheses and exploit possible correlations between them~\cite{PA-guedj19-10a}. We derive bounds of this type in Section~\ref{sec-a:uncond_pacb}.

Finally, we also consider the single-draw scenario. In this setting, we are interested in bounds of the following flavor: with probability at least $1-\delta$ under $P_{W\!\trainingdata}$,
\begin{equation}\label{B_eq:uncond_singledraw_generalization_bound_def}
\abs{\gen} \leq \epsilon.
\end{equation}
This type of result is relevant when we draw a single hypothesis $W$ based on our training data, and want to bound the generalization error of this particular $W$ with high probability. The probabilistic bounds in~\eqref{B_eq:uncond_pacb_generalization_bound_def} and~\eqref{B_eq:uncond_singledraw_generalization_bound_def} are said to be high-probability bounds if the dependence of $\epsilon$ on the confidence parameter $\delta$ is at most of order $\log(1/\delta)$. 

In Theorem~\ref{thm:unconditional_sub-Gaussian} below, we present exponential inequalities that will be used in Section~\ref{sec-a:uncond_exp}, Section~\ref{sec-a:uncond_pacb}, and Section~\ref{sec-a:uncond_sd} to derive generalization bounds of all three flavors. The derivation of these exponential inequalities and their use to obtain generalization bounds draw inspiration from \cite[Sec.~1.2.4]{PA-catoni07-a}, where a similar approach is used to obtain PAC-Bayesian and single-draw bounds for the special case in which the loss function has range restricted to $\{0,1\}$.
\begin{thm}\label{thm:unconditional_sub-Gaussian}
Let $\trainingdata=(Z_1,\dots,Z_n)\in\mathcal{Z}^n$ consist of $n$ \iid training samples generated from $P_Z$, and let $P_{W\vert \trainingdata}$ be a probabilistic learning algorithm.
Assume that $\ell(w,Z): \mathcal{W}\times \mathcal{Z}\rightarrow \reals$ is $\sigma$-sub-Gaussian under $P_Z$ for all $w\in \mathcal W$.
Also, assume that $P_{W\!\trainingdata}$ and~$P_W\!P_{\trainingdata}$ are mutually absolutely continuous with respect to each other. Then, for all~$\lambda\in \reals$,
\begin{equation}\label{B_eq:unconditional_sub-Gaussian_thm}
    \Exop_{P_{W\! \trainingdata}}\lefto[{\exp\lefto(\lambda \gen-\frac{\lambda^2\sigma^2}{2n}-\imath(W,\trainingdata)  \right)}\right]\leq 1.
\end{equation}
Furthermore,
\begin{equation}\label{B_eq:unconditional_sub-Gaussian_thm_for_tails}
    \Exop_{P_{W\!\trainingdata}}\!\biggo[\!\exp\biggo(\!\frac{n\!-\!1}{2\sigma^2}(\gen)^2\!-\!\log \!\sqrt{n}\! -\! \infdens\! \bigg)\!\bigg] \!\leq\! 1.\!
\end{equation}
\end{thm}
\begin{IEEEproof}
Since $\ell(w,Z)$ is $\sigma$-sub-Gaussian for all $w\in\mathcal{W}$ and the $Z_i$ are \iid, the training loss $(1/n)\sum_{i=1}^n  \ell(w,Z_i)$ is $\sigma/\sqrt{n}$-sub-Gaussian for all $w\in\mathcal{W}$, as remarked after Lemma~\ref{lem:hoeffdings_inequality}. Thus, for all $w\in\mathcal{W}$,
\begin{multline}\label{B_eq:proof_of_uncond_thm_sub-Gaussian_assuption}
    \Exop_{P_{\trainingdata} } \lefto[\exp\lefto(\lambda\left(\Exop_{P_Z}\lefto[\ell(w,Z)\right]-\frac{1}{n}\sum_{i=1}^n \ell(w,Z_i) \right)\right)\right]
   \\
   \leq \exp\lefto(\frac{\lambda^2\sigma^2}{2n}\right).
\end{multline}
Reorganizing terms and taking the expectation with respect to $P_W$, we get
\begin{equation}\label{B_eq:thm_1_proof_before_E}
    \Exop_{P_W\! P_{\trainingdata} } \lefto[\exp\lefto(\lambda \gen- \frac{\lambda^2\sigma^2}{2n}\right)\right]\leq 1.
\end{equation}
To obtain~\eqref{B_eq:unconditional_sub-Gaussian_thm}, we perform a change of measure from $P_W\! P_{\trainingdata}$  to $P_{W\! \trainingdata}$ \cite[Prop.~17.1]{PA-polyanskiy19-a}.

To obtain~\eqref{B_eq:unconditional_sub-Gaussian_thm_for_tails}, we note that~\eqref{B_eq:subgauss-square-lemma}, with~$\lambda=1-1/n$, implies that
\begin{equation}
\Ex{P_{\trainingdata} }{\exp\lefto(\frac{n-1}{2\sigma^2}(\genwZ)^2 \right) } \leq \sqrt{n}.
\end{equation}
We obtain the desired result by taking the expectation with respect to $P_W$, changing measure to $P_{W\!\trainingdata}$, and rearranging terms.

\end{IEEEproof}
Note that Theorem~\ref{thm:unconditional_sub-Gaussian} holds \emph{verbatim} if $P_W$ is replaced with an auxiliary distribution $Q_W$, under a suitable absolute continuity assumption. This is detailed in the next remark.
While an analogous statement holds for~\eqref{B_eq:unconditional_sub-Gaussian_thm_for_tails}, we present only the version similar to~\eqref{B_eq:unconditional_sub-Gaussian_thm} for brevity.
\begin{remark}\label{rem:uncond_conversion_to_Q_bounds}
Consider the setting of Theorem~\ref{thm:unconditional_sub-Gaussian}, but with the altered absolute continuity assumption that $P_{W\!\trainingdata}\ll Q_W\!P_{\trainingdata}$ and~$Q_W\!P_{\trainingdata} \ll P_{W\!\trainingdata}$ for some distribution $Q_W$ on $\mathcal{W}$. Then,
\begin{equation}\label{B_eq:unconditional_sub-Gaussian_with_Q}
    \Exop_{P_{W\! \trainingdata}}\lefto[{\exp\lefto(\lambda \gen\!-\!\frac{\lambda^2\sigma^2}{2n}\!-\!\log\frac{\dv P_{W\!\trainingdata} }{\dv Q_W\!P_{\trainingdata}}  \right)}\right]\!\leq\! 1 .\!
\end{equation}
\end{remark}
For the bounds that we will later derive, the choice $Q_W=P_W$ is optimal. Unfortunately, since the data distribution $P_Z$ is considered to be unknown in the statistical learning framework, the marginal distribution $P_W$ is also unavailable. Hence, $P_W$ needs to be replaced by some suitably chosen auxiliary distribution $Q_W$ whenever one wants to numerically evaluate the generalization bounds that we derive later in this section.\footnote{This issue is well understood in the PAC-Bayesian literature, where the available bounds are given in terms of an auxiliary (\emph{prior}) distribution $Q_W$ that does not depend on the unknown data distribution $P_Z$.} In the remainder of this paper, all bounds will be given in terms of $P_W$. Thanks to this choice, many of the terms that appear in our results will be expressible in terms of familiar information-theoretic quantities. However, through repeated references to Remark~\ref{rem:uncond_conversion_to_Q_bounds}, we will emphasize that the bounds can easily be generalized to the case in which $P_W$ is replaced by an auxiliary distribution $Q_W$.
\subsection{Average Generalization Error Bounds}\label{sec-a:uncond_exp}
We now use Theorem~\ref{thm:unconditional_sub-Gaussian} to obtain an average generalization error bound of the form given in~\eqref{B_eq:uncond_exp_generalization_bound_def}.
For this derivation, we use~\eqref{B_eq:unconditional_sub-Gaussian_thm} as our starting point rather than~\eqref{B_eq:unconditional_sub-Gaussian_thm_for_tails}.
As we later clarify, this leads to a more beneficial dependence on~$n$.

\begin{cor}\label{cor:exp_uncond}
Under the setting of Theorem~\ref{thm:unconditional_sub-Gaussian},
\begin{equation}\label{B_eq:cor_exp_uncond}
    \abs{ \Ex{P_{W\! \trainingdata}}{\gen}} \leq \sqrt{\frac{2\sigma^2}{n}I(W;\trainingdata)}.
\end{equation}
\end{cor}
\begin{IEEEproof}
We apply Jensen's inequality to~\eqref{B_eq:unconditional_sub-Gaussian_thm}, resulting in
\begin{equation}
    \exp\lefto(\!\!\lambda\Exop_{P_{W\! \trainingdata}}\lefto[\gen\right] \!-\! \frac{\lambda^2\sigma^2}{2n} 
 \!-\!\Ex{P_{W\! \trainingdata}}{\infdens}\! \right) \! \leq  \! 1.\!\!
\end{equation}
Note that $\Exop_{P_{W\! \trainingdata}}\lefto[\imath(W, \trainingdata)\right]=\relent{P_{W\!\trainingdata}}{P_W\!P_{\trainingdata} }=I(W;\trainingdata)$. By reorganizing terms, we get
\begin{equation}
     \lambda \Exop_{P_{W\! \trainingdata}}\lefto[\gen\right]-\lambda^2\frac{\sigma^2}{2n}  - \relent{P_{W\!\trainingdata}}{P_W\!P_{\trainingdata} } \leq 0.
\end{equation}
We now set $\lambda=n\Exop_{P_{W\! \trainingdata}}\lefto[\gen\right]/\sigma^2$ to optimize the bound, and thereby obtain
\begin{equation}\label{B_eq:exp_uncond_final_line_of_proof}
\Exop^2_{P_{W\! \trainingdata}}\lefto[\gen\right] - \frac{2\sigma^2}{n}\relent{P_{W\!\trainingdata}}{P_W\!P_{\trainingdata} }\leq 0
\end{equation}
from which~\eqref{B_eq:cor_exp_uncond} directly follows.
\end{IEEEproof}
The bound in~\eqref{B_eq:cor_exp_uncond} coincides with the result reported in~\cite[Thm.~1]{PA-xu17-05a}. As noted in Remark~\ref{rem:uncond_conversion_to_Q_bounds}, we can substitute an arbitrary $Q_W$ for $P_W$ in~\eqref{B_eq:exp_uncond_final_line_of_proof}, provided that the necessary absolute continuity criterion is fulfilled. This leads to a more general bound involving the relative entropy $\relent{P_{W\!\trainingdata}}{Q_W\!P_{\trainingdata} }$.
\subsection{PAC-Bayesian Generalization Error Bounds}\label{sec-a:uncond_pacb}
We now turn to PAC-Bayesian bounds of the form given in~\eqref{B_eq:uncond_pacb_generalization_bound_def}. In the following corollary, we reobtain a known data-dependent bound\footnote{An oft-cited source for this bound is~\cite{PA-mcallester98-07a}, although the bounds found therein differ slightly from what we present.} and present a novel data-independent version.
To derive the data-dependent bound, we need to use~\eqref{B_eq:unconditional_sub-Gaussian_thm_for_tails} as the starting point.
The reason for this is that, if we were to optimize the parameter~$\lambda$ in~\eqref{B_eq:unconditional_sub-Gaussian_thm}, we would incur a union bound cost.
However, when deriving the data-independent bound, we use~\eqref{B_eq:unconditional_sub-Gaussian_thm} as our starting point, similar to the average bound in~\eqref{B_eq:cor_exp_uncond}, as this avoids a logarithmic dependence on~$\sqrt n$.

\begin{cor}\label{cor:pacb_uncond}
Under the setting of Theorem~\ref{thm:unconditional_sub-Gaussian}, the following holds with probability at least $1-\delta$ under $P_{\trainingdata}$:
\begin{multline}\label{B_eq:cor_pacb_uncond}
\abs{ \Exop_{P_{W\vert \trainingdata} } \lefto[\gen\right] }  
\\
\leq \sqrt{\frac{2\sigma^2}{n-1}\left(\relent{P_{W\vert \trainingdata}}{P_W} +\log \frac{\sqrt{n}}{\delta}\right)}
\end{multline}
\begin{multline}
\label{B_eq:cor_pacb_uncond_second}
\abs{ \Exop_{P_{W\vert \trainingdata} } \lefto[\gen\right] }  \\ \leq  \sqrt{\frac{2\sigma^2}{n}\left(\frac
    {\Exop^{1/t}_{P_{\trainingdata}}\lefto[\relent{P_{W\vert \trainingdata}}{P_W}^t\right] }
    {(\delta/2)^{1/t}} +\log \frac{2}{\delta}\right)}
\end{multline}
for all $t>0$.
Here, the first inequality yields a data-dependent bound, while the second provides a data-independent version.
\end{cor}
\begin{IEEEproof}

We start from~\eqref{B_eq:unconditional_sub-Gaussian_thm_for_tails} and use Jensen's inequality, but only with respect to $P_{W\vert \trainingdata}$, unlike in the proof of Corollary~\ref{cor:exp_uncond}. This leads to
\begin{multline}\label{B_eq:pacb_uncond_after_jensen-data-dep}
\Exop_{P_{\trainingdata}}\!\lefto[\exp\lefto(\!\frac{n\!-\!1}{2\sigma^2}(\gen)^2\!-\!\log\! \sqrt{n}\!-\!\relent{P_{W\!\given\! \trainingdata}\!}{\!P_W}
\right) \right]\!\! \\
\leq 1
\end{multline}
where we used that, for a fixed $\trainingdata$,
\begin{equation}
\Ex{P_{W\vert \trainingdata}}{\imath(W,\trainingdata)}=\relent{P_{W\vert \trainingdata}}{P_W}.
\end{equation}
Next, we use the following result. Let $U\distas P_U$ be a nonnegative random variable satisfying $\Ex{}{U}\leq 1$. Then, Markov's inequality implies that
\begin{equation}\label{B_eq:markov_inequality_Pu}
P_U[U\leq 1/\delta]\geq 1- \Ex{ }{U}\delta\geq 1- \delta.
\end{equation}
By applying~\eqref{B_eq:markov_inequality_Pu} to the random variable in~\eqref{B_eq:pacb_uncond_after_jensen}, we obtain
\begin{multline}
P_{\trainingdata}\biggo[\exp\biggo(\frac{n\!-\!1}{2\sigma^2}\Ex{P_{W\vert \trainingdata} }{(\gen)^2}\!\\
-\log \sqrt{n}\!-\!\relent{P_{W\vert \trainingdata}}{P_W}
\bigg) \leq \frac{1}{\delta} 
\bigg] \geq  1-\delta.
\end{multline}
Reorganizing terms, we conclude that
\begin{multline}\label{B_eq:uncond_pacb_proof_markov_pz-data-dep}
    P_{\trainingdata}\biggo[\relent{P_{W\vert \trainingdata}}{P_W}-\frac{n-1}{2\sigma^2}\Ex{P_{W\vert \trainingdata} }{(\gen)^2} \\
    +\log\frac{\sqrt n}{\delta} \geq 0
\bigg] 
\geq 1-\delta.
\end{multline}
We obtain the desired result in~\eqref{B_eq:cor_pacb_uncond} by reorganizing terms and using Jensen's inequality.

We now turn to the data-independent bound.
As in Corollary~\ref{cor:exp_uncond}, we start from~\eqref{B_eq:unconditional_sub-Gaussian_thm} and use Jensen's inequality, but now only with respect to $P_{W\vert \trainingdata}$. This leads to
\begin{multline}\label{B_eq:pacb_uncond_after_jensen}
\Exop_{P_{\trainingdata}}\biggo[\exp\biggo(\lambda \Ex{P_{W\vert \trainingdata} }{\gen} - \frac{\lambda^2\sigma^2}{2n}\\
-\relent{P_{W\!\given\! \trainingdata}}{P_W}
\bigg) \bigg] \leq 1
\end{multline}
where we used that, for a fixed $\trainingdata$,
\begin{equation}
\Ex{P_{W\vert \trainingdata}}{\imath(W,\trainingdata)}=\relent{P_{W\vert \trainingdata}}{P_W}.
\end{equation}
By applying~\eqref{B_eq:markov_inequality_Pu} to the random variable in~\eqref{B_eq:pacb_uncond_after_jensen}, we obtain
\begin{multline}
P_{\trainingdata}\biggo[\exp\biggo(\lambda \Ex{P_{W\vert \trainingdata} } {\gen} - \frac{\lambda^2\sigma^2}{2n}\\
-\relent{P_{W\vert \trainingdata}}{P_W}
\bigg) \leq \frac{1}{\delta} 
\bigg] \geq  1-\delta.
\end{multline}
Reorganizing terms, we conclude that
\begin{multline}\label{B_eq:uncond_pacb_proof_markov_pz}
    P_{\trainingdata}\biggo[\frac{\lambda^2\sigma^2}{2n} -\lambda \Ex{P_{W\vert \trainingdata} } {\gen} 
    \\
    +\relent{P_{W\vert \trainingdata}}{P_W} +\log\frac{1}{\delta} \geq 0 
\bigg] \geq 1-\delta.
\end{multline}
We now apply Markov's inequality to the random variable $\relent{P_{W\vert \trainingdata}}{P_W}^t$, which after some manipulation yields
\begin{equation}\label{B_eq:uncond_pacb_proof_markov_with_t}
P_{\trainingdata}\!\lefto[\!\relent{P_{W\vert \trainingdata}\!}{\!P_W} \!\leq \!
\frac
    {\Exop^{1/t}_{P_{\trainingdata}}\lefto[\relent{P_{W\vert \trainingdata}\!}{\!P_W}^t\right] }
    {\delta^{1/t}} \right]
    \geq 1\!-\!\delta.\!
\end{equation}

Using~\eqref{B_eq:uncond_pacb_proof_markov_with_t} in~\eqref{B_eq:uncond_pacb_proof_markov_pz}, we obtain
\begin{multline} 
     P_{\trainingdata}\Biggl[ \frac{\lambda^2\sigma^2}{2n} -\lambda \Ex{P_{W\given \trainingdata}}{\gen} 
    \\
    +\frac{\Exop_{P_{\trainingdata}}^{1/t}\lefto[\relent{P_{W\given\trainingdata}}{P_W}^t\right]}{\delta^{1/t}}
     +\log \frac{1}{\delta}\geq 0\Biggr]\geq 1-2\delta.  \label{eq:combined_data_indep}
\end{multline}
Now, let $a={\Exop_{P_{\trainingdata}}^{1/t}\lefto[\relent{P_{W\given\trainingdata}}{P_W}^t\right]}/{\delta^{1/t}}
+\log \frac{1}{\delta}$ and~$b=\sigma^2/(2n)$.
The desired result in~\eqref{B_eq:cor_pacb_uncond_second} follows by setting $\lambda = \pm \sqrt{\frac{a}{b}}$ and performing the substitution $\delta\rightarrow \delta/2$.
\end{IEEEproof}
Note that by setting the parameter $t=1$ in~\eqref{B_eq:cor_pacb_uncond_second}, we get $\Exop_{P_{\trainingdata}}\lefto[\relent{P_{W\vert \trainingdata}}{P_W} \right]=I(W;\trainingdata)$. This choice recovers the result reported in~\cite[App.~3]{PA-bassily18-02a}. Instead, if we let $t\rightarrow \infty$, the polynomial $\delta$-dependence in~\eqref{B_eq:cor_pacb_uncond_second} disappears and the bound becomes a high-probability bound. This illustrates that one can get progressively better dependence on $\delta$ by letting the bound depend on higher moments of $\relent{P_{W\vert \trainingdata}}{P_W}$. The tightness of the resulting bound depends on how well one can control these higher moments.
Also, note that we can obtain a data-independent relaxation of~\eqref{B_eq:cor_pacb_uncond} by using~\eqref{B_eq:uncond_pacb_proof_markov_with_t} in~\eqref{B_eq:cor_pacb_uncond}.
However, the resulting bound would include a~$2\sigma^2\log(\sqrt n)/(n-1)$ term in the square root, which is not present in~\eqref{B_eq:cor_pacb_uncond_second}.
Finally, as per Remark~\ref{rem:uncond_conversion_to_Q_bounds}, we can obtain more general bounds by replacing $P_W$ in~\eqref{B_eq:cor_pacb_uncond} and~\eqref{B_eq:cor_pacb_uncond_second} with an arbitrary $Q_W$ that satisfies a suitable absolute continuity property.

\subsection{Single-Draw Generalization Error Bounds}\label{sec-a:uncond_sd}
We now turn our attention to single-draw bounds of the form given in~\eqref{B_eq:uncond_singledraw_generalization_bound_def}.
We will derive generalization bounds by using two different approaches. 
Our first approach relies on the exponential inequalities in Theorem~\ref{thm:unconditional_sub-Gaussian}.
First, we use~\eqref{B_eq:unconditional_sub-Gaussian_thm_for_tails} to get a data-dependent bound in terms of the information density $\infdens$.
We then obtain several data-independent bounds on the basis of~\eqref{B_eq:unconditional_sub-Gaussian_thm}.
Our second approach, which yields a generalization bound that is explicit in the tail of the information density, relies on the change of measure result stated in Lemma~\ref{lem:strong_converse_lemma}. This bound can be relaxed to obtain essentially the same data-independent bounds obtained using the first approach.

\subsubsection{Generalization Bounds from the Exponential Inequalities}
We begin by using Theorem~\ref{thm:unconditional_sub-Gaussian} to derive a data-dependent single-draw generalization bound and a data-independent version, similar to the PAC-Bayesian results in Corollary~\ref{cor:pacb_uncond}. Both of these bounds are novel.
\begin{cor}\label{cor:singledraw_uncond}
Under the setting of Theorem~\ref{thm:unconditional_sub-Gaussian}, with probability at least $1-\delta$ under $P_{W\!\trainingdata}$, the following inequalities hold for all $t>0$:\footnote{Note that the argument of the square root in~\eqref{B_eq:cor_singledraw_uncond} can be negative, but that this happens with probability at most $\delta$. Therefore, the right-hand side of~\eqref{B_eq:cor_singledraw_uncond} is well-defined with probability at least $1-\delta$.}
\begin{align}\label{B_eq:cor_singledraw_uncond}
  \abs{ \gen } &\leq \sqrt{{\frac{2\sigma^2}{n-1}\left(\imath(W,\trainingdata) +\log \frac{\sqrt{n}}{\delta}\right)}}\\\label{B_eq:cor_singledraw_uncond_second}
   \abs{ \gen } & \!\leq  \! \sqrt{\frac{2\sigma^2}{n}\!\left(\!I(W;\!\trainingdata) \!+\! \frac{M_t(W;\trainingdata)}{(\delta/2)^{1/t}} \!+\!\log\! \frac{2}{\delta}\!\right)}.\!
\end{align}
Here, the first inequality provides a data-dependent bound and the second inequality is a data-independent version. In~\eqref{B_eq:cor_singledraw_uncond_second}, $M_t(W;\trainingdata)$ is the $t$th root of the $t$th central moment of $\imath(W,\trainingdata)$:
\begin{equation}\label{B_eq:def_M_P_WZn}
M_t(W;\trainingdata) = \Exop^{1/t}_{P_{W\! \trainingdata}}\lefto[\abs{ \imath(W,\trainingdata) - \relent{P_{W\!\trainingdata}}{P_W\!P_{\trainingdata} }}^t \right].
\end{equation}
\end{cor}
\begin{IEEEproof}
By directly applying Markov's inequality~\eqref{B_eq:markov_inequality_Pu} to~\eqref{B_eq:unconditional_sub-Gaussian_thm_for_tails}, we conclude that
\begin{multline}
P_{W\! \trainingdata}\lefto[\exp\biggo(\frac{n\!-\!1}{2\sigma^2}(\gen)^2\!-\!\log \sqrt{n} \!-\! \infdens\! \right)\leq \frac{1}{\delta} \bigg] \\
\geq 1-\delta
\end{multline}
from which~\eqref{B_eq:cor_singledraw_uncond} follows after reorganizing terms.
We now turn to the data-independent version.
By directly applying Markov's inequality~\eqref{B_eq:markov_inequality_Pu} to~\eqref{B_eq:unconditional_sub-Gaussian_thm}, we conclude that
\begin{multline}\label{B_eq:singledraw_uncond_dataindep_before_relax}
P_{W\! \trainingdata}\lefto[\exp\biggo(\lambda \gen - \frac{\lambda^2\sigma^2 }{2n}
-\imath(W,\trainingdata) \right)\leq \frac{1}{\delta} \bigg] \\
\geq 1-\delta.
\end{multline}
We now use Markov's inequality in the following form: for a random variable $U\distas P_U$, the following holds for all $t>0$:
\begin{equation}\label{B_eq:markov_for_arbitrary_u}
P_U\lefto[U \leq \Exop[U]+ \frac{\Exop^{1/t}[\abs{U-\Exop[U]}^t]}{\delta^{1/t}} \right]\geq 1-\delta.
\end{equation}
Applying~\eqref{B_eq:markov_for_arbitrary_u} with $U=\imath(W,\trainingdata)$ and using the union bound to combine the resulting inequality with~\eqref{B_eq:cor_singledraw_uncond}, we obtain
\begin{multline}
P_{W\! \trainingdata}\lefto[\exp\biggo(\lambda \gen - \frac{\lambda^2\sigma^2 }{2n}
-\frac{M_t(W;\trainingdata)}{\delta^{1/t}}) \right)\leq \frac{1}{\delta} \bigg] \\
\geq 1-2\delta.
\end{multline}
Let $a=\frac{M_t(W;\trainingdata)}{\delta^{1/t}}
+\log \frac{1}{\delta}$ and $b=\sigma^2/(2n)$.
The desired result in~\eqref{B_eq:cor_singledraw_uncond_second} now follows by setting $\lambda = \pm \sqrt{\frac{a}{b}}$ and performing the substitution $\delta \rightarrow \delta/2$.
\end{IEEEproof}
As usual, we can obtain more general bounds by substituting $Q_W$ for $P_W$ in Corollary~\ref{cor:singledraw_uncond}, provided that the necessary absolute continuity assumption is satisfied.
Again, we can obtain a data-independent relaxation of~\eqref{B_eq:cor_singledraw_uncond} by directly applying~\eqref{B_eq:markov_for_arbitrary_u} with~$U=\imath(W,\trainingdata)$.
However, as before, this would lead to an extra logarithmic dependence on~$\sqrt n$ compared to~\eqref{B_eq:cor_singledraw_uncond_second}.

Similarly to what we noted for the PAC-Bayesian bound~\eqref{B_eq:cor_pacb_uncond_second}, the $\delta$-dependence in~\eqref{B_eq:cor_singledraw_uncond_second} can be made more benign by letting the bound depend on higher central moments of $\infdens$, but the tightness of the resulting bound hinges on how well one can control these higher moments. In particular, if we let $t\rightarrow \infty$ in~\eqref{B_eq:cor_singledraw_uncond_second}, we obtain the following high-probability bound:
 \begin{equation}\label{B_eq:cor_singledraw_uncond_t_inf}
    \abs{  \gen } \!\leq  \!
     \sqrt{\frac{2\sigma^2}{n}\!\left(\!I(W;\!\trainingdata)\!+\!{M_\infty(W;\!\trainingdata)} \!+\!\log\! \frac{2}{\delta}\right)}.\!
\end{equation}
Here, ${M_\infty(W;\trainingdata)}$ is given by
\begin{equation}\label{B_eq:def_M_infinity}
 M_\infty(W;\trainingdata)=\esssup_{P_{W\!\trainingdata}} \abs{\imath(W,\trainingdata)- I(W;\trainingdata)}.
\end{equation}
Note that the supremization in~\eqref{B_eq:def_M_infinity} is over the argument of $\infdens$, whereas $I(W;\trainingdata)$ is a constant.

The data-independent version in Corollary~\ref{cor:singledraw_uncond} is not as tight as the one obtained in Corollary~\ref{cor:pacb_uncond}. Indeed, since $\infdens$ can be negative, we had to use a weaker version of Markov's inequality (compare~\eqref{B_eq:markov_for_arbitrary_u} with~\eqref{B_eq:markov_inequality_Pu}). In the following corollary, we provide two alternative data-independent bounds. The first bound depends on the maximal leakage $\mathcal{L}(\trainingdata\rightarrow W)$ defined in~\eqref{B_eq:def_maximal_leakage}, and recovers~\cite[Cor.~10]{PA-esposito19-12a} up to a logarithmic term. The second bound, which is novel, is in terms of the R\'enyi divergence~\eqref{B_eq:def_renyi_divergence}.
\begin{cor}\label{cor:singledraw_uncond_alphadiv_leakage}
Under the setting of Theorem~\ref{thm:unconditional_sub-Gaussian}, the following inequalities hold with probability at least $1-\delta$ under $P_{W\!\trainingdata}$:
\begin{equation}\label{B_eq:cor_singledraw_uncond_leakage}
\abs{ \gen } \leq \sqrt{\frac{2\sigma^2}{n}\left(\mathcal{L}(\trainingdata\rightarrow W) +2\log \frac{2}{\delta}\right)}
\end{equation}
and, for all $\alpha,\gamma> 1$ such that $1/\alpha+1/\gamma=1$,
\begin{multline}\label{B_eq:cor_singledraw_uncond_alphadiv}
\abs{ \gen }\! \leq \! \bigg(\frac{2\sigma^2}{n}\!\!\bigg(\!\frac{\alpha\!-\!1}{\alpha} \alpharelent{\alpha}{P_{W\!\trainingdata}\!}{\!P_W\!P_{\trainingdata} } \\
+\!\frac{\gamma\!-\!1}{\gamma} \alpharelent{\gamma}{P_{W\!\trainingdata}\!}{\!P_W\!P_{\trainingdata}} \!+\!2\log\! \frac{2}{\delta}\bigg)\bigg)^{1/2}.\!
\end{multline}
\end{cor}
\begin{IEEEproof}
By applying Markov's inequality, we conclude that with probability at least $1-\delta$ under $P_{W\!\trainingdata}$,
\begin{equation}\label{B_eq:cor_7_proof_step_1}
\infdens \leq \log \Ex{P_{W\!\trainingdata}}{\frac{\dv P_{W\! \trainingdata}}{\dv P_W\!P_{\trainingdata}}} + \log\lefto(\frac{1}{\delta}\right).
\end{equation}
Next, the expectation over $P_{\trainingdata\vert W}$ can be replaced by an essential supremum to obtain the inequality
\begin{align}
\Ex{P_W\!P_{\trainingdata\vert W}}{\frac{\dv P_{W\! \trainingdata}}{\dv P_W\!P_{\trainingdata}}} \leq \Ex{P_W}{\esssup_{P_{\trainingdata\vert W}} \frac{\dv P_{W\! \trainingdata}}{\dv P_W\!P_{\trainingdata}}}.
\end{align}
The assumption that $P_{W\!\trainingdata}\ll P_W\!P_{\trainingdata}$ means that any set in the support of $P_{W\!\trainingdata}$ is also in the support of $P_W\!P_{\trainingdata}$. We can therefore upper-bound the $\esssup$ as
\begin{align}\label{B_eq:cor_7_proof_step_3}
\esssup_{P_{\trainingdata\vert W}} \frac{\dv P_{W\! \trainingdata}}{\dv P_W\!P_{\trainingdata}} \leq \esssup_{P_{\trainingdata}} \frac{\dv P_{W\! \trainingdata}}{\dv P_W\!P_{\trainingdata}}.
\end{align}
By using the union bound to combine~\eqref{B_eq:cor_7_proof_step_1}-\eqref{B_eq:cor_7_proof_step_3} with~\eqref{B_eq:singledraw_uncond_dataindep_before_relax} and following steps analogous to the proof of~\eqref{B_eq:cor_singledraw_uncond_second}, we obtain~\eqref{B_eq:cor_singledraw_uncond_leakage}.

To prove~\eqref{B_eq:cor_singledraw_uncond_alphadiv}, we first apply Markov's inequality and then perform a change of measure to conclude that the following inequalities hold with probability at least $1-\delta$ under $P_{W\! \trainingdata}$:
\begin{align}\label{B_eq:cor_singledraw_uncond_alphadiv_pf_markov_com}
\imath(W,\trainingdata) &\leq \log\Exop_{P_{W\!\trainingdata}}\lefto[\frac{\dv P_{W\! \trainingdata}}{\dv P_W\!P_{\trainingdata}}\right] + \log\frac{1}{\delta} \\
&\leq  \log\Exop_{P_W\!P_{\trainingdata}}\lefto[\left(\frac{\dv P_{W\! \trainingdata}}{\dv P_W\!P_{\trainingdata}}\right)^2\right] + \log\frac{1}{\delta}.
\end{align}
Next, we apply H\"older's inequality twice as follows. Let $\alpha,\gamma,\alpha',\gamma'> 1$ be constants such that $1/\alpha+1/\gamma=1/\alpha'+1/\gamma'=1$. Then,
\begin{align}
&\Exop_{P_W\!P_{\trainingdata}}\lefto[\left(\frac{\dv P_{W\! \trainingdata}}{\dv P_W\!P_{\trainingdata}}\right)^2\right] \nonumber
\\
& \!\leq \!  \Exop_{P_W}\lefto[\Exop_{P_{\trainingdata}}^{1/\alpha}\lefto[e^{\alpha\infdens}\right]\cdot\Exop_{P_{\trainingdata}}^{1/\gamma}\lefto[e^{\gamma\infdens}\right]\right] \\
    & \!\leq\!   \Exop_{P_W}^{1/\alpha'}\lefto[\Exop_{P_{\trainingdata}}^{\alpha'/\alpha}\lefto[e^{\alpha\infdens}\right]\right]\!\cdot\!\Exop_{P_W}^{1/\gamma'}\lefto[\Exop_{P_{\trainingdata}}^{\gamma'/\gamma}\lefto[e^{\gamma\infdens}\right]\right]\!. \!
\end{align}
Setting $\alpha=\alpha'$, which implies $\gamma=\gamma'$, we conclude that
\begin{align}
 &\log\Exop_{P_W\!P_{\trainingdata}}\lefto[\left(\frac{\dv P_{W\! \trainingdata}}{\dv P_W\!P_{\trainingdata}}\right)^2\right] \nonumber\\
& \leq  \log \Exop_{P_W\!P_{\trainingdata} }^{1/\alpha}\lefto[e^{\alpha\infdens}\right] + \log \Exop_{P_W\!P_{\trainingdata} }^{1/ \gamma}\lefto[e^{\gamma\infdens}\right] \label{B_eq:choosePandQ} \\ \label{B_eq:cor_uncond_sd_alphadiv_proof_infdens_bound}
    &=\!\frac{\alpha\!-\!1}{\alpha}D_\alpha(P_{W\! \trainingdata}\!\,||\,\!P_W\!P_{\trainingdata})\!+\!\frac{\gamma\!-\!1}{\gamma}D_\gamma(P_{W\! \trainingdata}\!\,||\,\!P_W\!P_{\trainingdata}). \!
\end{align}
Substituting~\eqref{B_eq:cor_uncond_sd_alphadiv_proof_infdens_bound} into~\eqref{B_eq:cor_singledraw_uncond_alphadiv_pf_markov_com}, and then combining~\eqref{B_eq:cor_singledraw_uncond_alphadiv_pf_markov_com} with~\eqref{B_eq:singledraw_uncond_dataindep_before_relax} through the union bound, we establish~\eqref{B_eq:cor_singledraw_uncond_alphadiv} after following steps analogous to the proof of~\eqref{B_eq:cor_singledraw_uncond_second}.
\end{IEEEproof}
The bound in~\eqref{B_eq:cor_singledraw_uncond_leakage} coincides with the maximal leakage bound in~\cite[Cor.~10]{PA-esposito19-12a}, up to a $2\sigma^2\log(2/\delta)/n$ term inside the square root. It is stronger than the max information bound in~\cite[Thm.~4]{PA-dwork15-06a}, for the case in which the parameter $\beta$ therein is set to $0$, and also stronger than~\eqref{B_eq:cor_singledraw_uncond_t_inf}, up to the same logarithmic term.  Indeed, let the max information be defined as
\begin{equation}\label{B_eq:max_mi}
    I_{\textnormal{max}}(W;\trainingdata) = \esssup_{P_{W\!\trainingdata} }\imath(W,\trainingdata).
\end{equation}
As shown in~\cite[Lem.~12]{PA-esposito19-12a}, $\mathcal{L}(\trainingdata \rightarrow W)\leq I_{\textnormal{max}}(W;\trainingdata)$. It is also readily verified that
\begin{equation}I_{\textnormal{max}}(W;\trainingdata)\leq I(W;\trainingdata)+{M_\infty(W;\trainingdata)}.
\end{equation}
We thus have the chain of inequalities
\begin{equation}
\mathcal{L}(\trainingdata\rightarrow W) \leq  I_{\textnormal{max}}(W;\trainingdata) \leq I(W;\trainingdata)+{M_\infty(W;\trainingdata)}.
\end{equation}
In particular, provided that
\begin{equation} \mathcal{L}(\trainingdata \rightarrow W) \leq I_{\textnormal{max}}(W;\trainingdata) + \log\frac{2}{\delta}
\end{equation}
the bound in~\eqref{B_eq:cor_singledraw_uncond_leakage} is tighter than the max information bound in~\cite[Thm.~4]{PA-dwork15-06a} with $\beta=0$, and also tighter than~\eqref{B_eq:cor_singledraw_uncond_t_inf}. Still, the bound in~\cite[Cor.~10]{PA-esposito19-12a} is stronger due to the aforementioned logarithmic term.

As usual, we can obtain more general bounds by replacing $P_W$ with an arbitrary $Q_W$ in Corollary~\ref{cor:singledraw_uncond_alphadiv_leakage}, provided that $P_{W\!\trainingdata}\ll Q_W\!P_{\trainingdata}$. However, for the proof of~\eqref{B_eq:cor_singledraw_uncond_leakage}, we still need the original absolute continuity assumption $P_{W\!\trainingdata}\ll P_W\!P_{\trainingdata}$ to guarantee that~\eqref{B_eq:cor_7_proof_step_3} holds. Note that a similar extension can easily be performed on~\cite[Thm.~1]{PA-esposito19-12a} and on the corollaries that are based on it, including~\cite[Cor. 10]{PA-esposito19-12a}.

\subsubsection{Generalization Bounds from the Strong Converse}\label{sec-a:uncond_sd_strong_conv}
Next, we use Lemma~\ref{lem:strong_converse_lemma} to derive an additional data-independent single-draw generalization bound. This novel bound depends on the tail of the information density.
\begin{thm}\label{thm:unconditional_hypothesis_main_bound}
Under the setting of Theorem~\ref{thm:unconditional_sub-Gaussian}, with probability at least $1-\delta$ under $P_{W\! \trainingdata}$, the following holds:
\begin{multline}\label{B_eq:thm_unconditional_hypothesis_main_bound}
    \abs{\gen}\\
    \leq\sqrt{ \frac{2\sigma^2}{n}
    \left(
        \gamma + \log\lefto(\frac{2}{\delta-P_{W\! \trainingdata}\lefto[\imath(W,\trainingdata)\geq \gamma\right]} \right)
    \right)}.
\end{multline}
This is valid for all $\gamma$ such that the right-hand side is defined and real.
\end{thm}
\begin{IEEEproof}
The proof relies on Lemma~\ref{lem:strong_converse_lemma}. We set $P=P_{W\!\trainingdata}$, $Q=P_W\!P_{\trainingdata}$, and%
\begin{equation}\label{B_eq:high_error_event}
\setE=\{W,\trainingdata: \abs{\gen}>\epsilon \}.
\end{equation}
Due to the $\sigma$-sub-Gaussianity of the loss function, Lemma~\ref{lem:hoeffdings_inequality} implies that
\begin{align}\label{B_eq:thm_main_uncond_hyp_bound_pf_QE_bound}
P_W\!P_{\trainingdata}[\setE] &=P_W\!P_{\trainingdata}\lefto[\abs{ L_{\trainingdata}(W)-\Exop_{P_{\trainingdata} }\lefto[L_{\trainingdata}(W)\right]}>\epsilon  \right] \\
&\leq 2\exp\lefto(-\frac{n\epsilon^2}{2\sigma^2} \right).
\end{align}
Substituting~\eqref{B_eq:thm_main_uncond_hyp_bound_pf_QE_bound} into~\eqref{B_eq:lem_strong_converse_lemma}, we get
\begin{multline}\label{B_eq:strong_conv_solve_for_eps}
    P_{W\! \trainingdata}[\abs{\gen}>\epsilon] \\
    \leq P_{W\! \trainingdata}\lefto[\imath(W,\trainingdata)\geq \gamma\right] + 2\exp\lefto(\gamma-n\frac{\epsilon^2}{2\sigma^2}\right).
\end{multline}
We obtain the desired result by requiring the right-hand side of~\eqref{B_eq:strong_conv_solve_for_eps} to equal $\delta$ and solving for $\epsilon$.
\end{IEEEproof}
As for the previous results, a more general bound can be obtained by setting $Q=Q_W\!P_{\trainingdata}$, where $Q_W$ is an arbitrary auxiliary distribution on $\mathcal{W}$, provided that a suitable absolute continuity criterion is fulfilled.

The result in Theorem~\ref{thm:unconditional_hypothesis_main_bound} indicates a trade-off between the decay of the tail of the information density and the tightness of the generalization bound. Indeed, the parameter $\gamma$ has to be chosen sufficiently large to make the argument of the logarithm positive. However, increasing $\gamma$ too much may yield a loose bound because of the $\gamma$ term that is added to the logarithm.

The bound in Theorem~\ref{thm:unconditional_hypothesis_main_bound} can be relaxed to recover some of the data-independent bounds discussed earlier in this section, up to a $2\sigma^2\log(2)/n$ penalty term inside the square root. In Remarks~\ref{rem:uncond_alt_deriv_moment} and~\ref{rem:uncond_alt_deriv_leakage}, we present these alternative derivations.
\begin{remark}[Alternative derivation of the moment bound{~\eqref{B_eq:cor_singledraw_uncond_second}}]\label{rem:uncond_alt_deriv_moment}
Using Markov's inequality, we conclude that
\begin{align}
&P_{W\! \trainingdata}\lefto[\imath(W,\!\trainingdata)\geq \gamma\right]\nonumber \\
&\leq\!   P_{W\! \trainingdata}\!\biggo[\!\!\abs{\imath(W,\!\trainingdata)\!-\!\relent{P_{W\!\trainingdata}\!}{\!P_W\!P_{\trainingdata} }}\!\geq\! \gamma\!-\!\relent{P_{W\!\trainingdata}\!}{\!P_W\!P_{\trainingdata} }\!\bigg]\!\!\nonumber\\
 &\leq \frac{(M_t(W;\trainingdata))^t }{(\gamma-\relent{P_{W\!\trainingdata}}{P_W\!P_{\trainingdata} })^{t}}\label{B_eq:moment_bound_from_hyptest_markov}
\end{align}
where $M_t(W;\trainingdata)$ is defined in~\eqref{B_eq:def_M_P_WZn}. Next, we set
\begin{equation}%
    \gamma =\relent{P_{W\!\trainingdata}}{P_W\!P_{\trainingdata}}+\frac{M_t(W;\trainingdata)}{(\delta/2)^{1/t}}
\end{equation}
which, once it is substituted into~\eqref{B_eq:moment_bound_from_hyptest_markov}, implies that we have $P_{W\! \trainingdata}\lefto[\imath(W,\trainingdata)\geq \gamma\right] \leq {\delta}/{2}$. Using this inequality in~\eqref{B_eq:thm_unconditional_hypothesis_main_bound}, we obtain
\begin{multline}
    \abs{\gen}
   \\
   \leq\sqrt{ \frac{2\sigma^2}{n}\!
    \left(\!
        \relent{P_{W\!\trainingdata}}{P_W\!P_{\trainingdata}}\!+\!\frac{M_t(W;\trainingdata)}{(\delta/2)^{1/t}} \!+\! \log\!\frac{4}{\delta}
    \right)}.
\end{multline}
This coincides with the bound in~\eqref{B_eq:cor_singledraw_uncond_second}, up to a $2\sigma^2\log(2)/n$ term inside the square root.
\end{remark}
\begin{remark}[Alternative derivation of the maximal leakage bound~\eqref{B_eq:cor_singledraw_uncond_leakage}]\label{rem:uncond_alt_deriv_leakage}
Note that
\begin{equation}
P_{W\! \trainingdata}[\imath(W,\trainingdata) \!\geq\! \gamma] \!\leq\! P_W\lefto[\esssup_{P_{\trainingdata\vert W}} \exp\lefto(\infdens \right)\geq e^\gamma\right].\!
\end{equation}
Since $P_{W\!\trainingdata}\ll P_W\!P_{\trainingdata}$, the $\esssup$ can be upper-bounded as in~\eqref{B_eq:cor_7_proof_step_3}. Hence,
\begin{equation}\label{B_eq:max_leak_alt_deriv_esssup_weakening}
P_{W\! \trainingdata}[\imath(W,\trainingdata) \!\geq\! \gamma] \!\leq\! P_W\lefto[\esssup_{P_{\trainingdata}}\exp\lefto(\infdens \right)\!\geq \!e^\gamma\right].\!
\end{equation}
By applying Markov's inequality to the right-hand side of~\eqref{B_eq:max_leak_alt_deriv_esssup_weakening}, we find that
\begin{align}\label{B_eq:max_leak_alt_deriv_tail_bound}
P_{W\! \trainingdata}[\imath(W,\trainingdata)\! \geq\! \gamma] &\!\leq\! e^{-\gamma}\Ex{P_W}{\esssup_{P_{\trainingdata}}\exp\lefto(\infdens \right)}\!\\
&=e^{-\gamma} \exp\lefto(\mathcal{L}(\trainingdata \rightarrow W)\right).
\end{align}
Substituting~\eqref{B_eq:max_leak_alt_deriv_tail_bound} into~\eqref{B_eq:thm_unconditional_hypothesis_main_bound} and setting $\gamma = \mathcal{L}(\trainingdata \rightarrow W) + \log(2/\delta)$, we conclude that with probability at least $1-\delta$ under $P_{W\! \trainingdata}$,
\begin{equation}
\abs{\gen}\!\leq\! \sqrt{ \frac{2\sigma^2}{n}\left(\mathcal{L}(\trainingdata\rightarrow W)\!+\!\log 2\!+\!2\log\! \frac{2}{\delta} \right) }.\!
\end{equation}
This coincides with the maximal leakage bound in~\eqref{B_eq:cor_singledraw_uncond_leakage} up to a $2\sigma^2\log(2)/n$ term inside the square root, and with~\cite[Cor.~10]{PA-esposito19-12a} up to a $2\sigma^2\log (4/\delta)/n$ term inside the square root.
\end{remark}
\section{Generalization Bounds for the CMI setting}
We now consider the CMI setting described in Section~\ref{sec-a:preliminaries}. For this setting, we will require the stronger assumption that the loss function $\ell(\cdot,\cdot)$ is bounded, rather than the sub-Gaussian assumption in Section~\ref{sec-a:uncond_results}.
As detailed in the proof of Theorem~\ref{thm:conditional_bounded_inequality} below, boundedness will be crucial to establish inequalities similar to~\eqref{B_eq:unconditional_sub-Gaussian_thm} and~\eqref{B_eq:unconditional_sub-Gaussian_thm_for_tails} for the case in which the expectation over $\supersample$ is replaced by an expectation over the selection random variable $\subsetchoice$.

The bounds in this section will depend on the conditional information density~\eqref{B_eq:conditional_info_density}. Intuitively, rather than asking how much information on the training data $\trainingdata$ can be inferred from $W$, we instead ask how much information $W$ reveals about whether $\tilde Z_{i}$ or $\tilde Z_{i+n}$ has been used for training, for $i=1,\dots,n$, given the knowledge of $\supersample$. We will make this intuition more precise and highlight the advantages of the random-subset approach when we compare the generalization error bounds obtained in this section to the ones in Section~\ref{sec-a:uncond_results}, under the assumption of a bounded loss function.

As in Section~\ref{sec-a:uncond_results}, the generalization bounds in this section will take different forms: average generalization bounds, PAC-Bayesian bounds, and single-draw bounds. The average bound for the CMI setting has a form similar to~\eqref{B_eq:uncond_exp_generalization_bound_def}, namely
\begin{equation}\label{B_eq:conditional_exp_gen_error_def}
\abs{\Exop_{P_{W\!\supersample \!\subsetchoice} }\lefto[ \textnormal{gen}(W,\trainingdata(\subsetchoice)) \right] } \leq \epsilon.
\end{equation}
For the PAC-Bayesian and single-draw settings, it will turn out to be convenient to first obtain probabilistic bounds on the following quantity:
\begin{equation}\label{B_eq:genshat_definition}
\genShat \!=\! \frac{1}{n}\sum_{i=1}^n \lefto(\ell(W, Z_i(\bar S_i)) \!-\! \ell(W,Z_i(S_i)) \right).
\end{equation}
Here, $\bar \subsetchoice$ is a vector whose entries are modulo-$2$ complements of the entries of $\subsetchoice$. As a consequence, $\trainingdata(\bar\subsetchoice)$ contains all the elements of $\supersample$ that are not in $\trainingdata(\subsetchoice)$. So, instead of comparing the loss on the training data to the expected loss on a new sample, we compare it to a test loss, i.e., the loss on $n$ samples that are independent of $W$. Note that quantities similar to~\eqref{B_eq:genshat_definition} are what one computes when empirically assessing the generalization performance of a learning algorithm.

In the PAC-Bayesian setting, we will be interested in deriving bounds of the following form: with probability at least $1-\delta$ under $P_{\supersample\!\subsetchoice} =P_{\supersample}P_{\subsetchoice}$,
\begin{equation}\label{B_eq:conditional_pacb_gen_error_def}
\abs{ \Exop_{P_{W\vert \supersample\!\subsetchoice}} \lefto[\genShat \right] } \leq \epsilon.
\end{equation}
Similarly, in the single-draw setting, the bounds of interest will be of the following form: with probability at least $1-\delta$ under $P_{W\!\supersample\!\subsetchoice}=P_{W\vert \supersample\!\subsetchoice}P_{\supersample}P_{\subsetchoice}$,
\begin{equation}\label{B_eq:conditional_sd_gen_error_def}
\abs{ \genShat } \leq \epsilon.
\end{equation}
As we establish in Theorem~\ref{thm:hoeffding_genshat_to_gens} below, the probabilistic bounds on $\genShat$ given in~\eqref{B_eq:conditional_pacb_gen_error_def} and~\eqref{B_eq:conditional_sd_gen_error_def} can be converted into probabilistic bounds on $\genS$ by adding a $\delta$-dependent penalty term.
\begin{thm}\label{thm:hoeffding_genshat_to_gens}
Let $\supersample=(\tilde Z_1,\dots,\tilde Z_{2n})\in \mathcal{Z}^{2 n}$ consist of $2n$ \iid training samples generated from $P_Z$ and let $\subsetchoice$ be a random vector, independent of $\supersample$, with entries drawn independently from a $\mathrm{Bern}(1/2)$ distribution. Let $\trainingdata(\subsetchoice)$ denote the subset of $\supersample$~obtained through $\subsetchoice$ by the rule $Z_i(S_i)=\tilde Z_{i+S_in}$, for $i=1,\dots,n$. Also, let $\bar \subsetchoice$ be the modulo-$2$ complement of $\subsetchoice$. Let $P_{W\vert \trainingdata(\subsetchoice)}$ be a randomized learning algorithm.\footnote{Note that, by construction, $W$ and $(\supersample,\subsetchoice)$ are conditionally independent given $\trainingdata(\subsetchoice)$.} Assume that $\ell(w,z)$ is bounded on $[a,b]$ for all $w\in \mathcal W$ and all $z\in\mathcal{Z}$. Also, assume that the following two probabilistic inequalities hold: with probability at least $1-\delta$ under $P_{W\!\supersample\!\subsetchoice}$,
\begin{equation}\label{B_eq:pf_of_lem_genShat_condition}
\abs{\genShat} \leq \epsilon_{\textnormal{SD}}(\delta)
\end{equation}
and with probability at least $1-\delta$ under $P_{\supersample\! \subsetchoice}$,
\begin{equation}\label{B_eq:pf_of_lem_genShat_condition_pb}
\abs{\Exop_{P_{W\vert \supersample\!\subsetchoice}}\lefto[\genShat\right]} \leq \epsilon_{\textnormal{PB}}(\delta).
\end{equation}
Then, with probability at least $1-\delta$ under $P_{W\!\supersample\!\subsetchoice}$,
\begin{equation}\label{B_eq:lem_genShat_to_gen_bound}
\abs{ \textnormal{gen}(W,\trainingdata(\subsetchoice)) } \leq \epsilon_{\textnormal{SD}}\lefto(\frac{\delta}{2}\right) +  \sqrt{\frac{(b-a)^2}{2n}\log\frac{4}{\delta}}
\end{equation}
and with probability at least $1-\delta$ under $P_{\supersample\! \subsetchoice}$,
\begin{equation}\label{B_eq:lem_genShat_to_gen_bound_pb}
\abs{\Exop_{P_{W\vert \supersample\!\subsetchoice}}\lefto[\textnormal{gen}(W,\!\trainingdata(\subsetchoice))\right]}\! \leq\! \epsilon_{\textnormal{PB}}\lefto(\!\frac{\delta}{2}\!\right) \!+ \! \sqrt{\!\frac{(b\!-\!a)^2}{2n}\log\!\frac{4}{\delta}}.\!
\end{equation}
\end{thm}
\begin{IEEEproof}
Since $\ell(w,Z_i(S_i))$ is bounded on $[a,b]$ for all $i=1,\dots,n$, it is $(b-a)/2$-sub-Gaussian for all $w\in \mathcal{W}$. From this, it follows that $L_{\trainingdata(\bar\subsetchoice)}(w)$ is $(b-a)/(2\sqrt{n})$-sub-Gaussian for all $w\in \mathcal{W}$. Hence, using Lemma~\ref{lem:hoeffdings_inequality}, we have that, for all $\epsilon>0$,
\begin{align}
&\abs{ L_{\trainingdata(\bar\subsetchoice)}(W)\! -\! \Exop_{P_Z}\lefto[\ell(W,Z)\right] } \nonumber\\
&=\!\abs{ \frac{1}{n}\!\sum_{i=1}^n\!\ell\lefto(W,Z_i(\bar S_i)\right)\! -\! \Exop_{P_{\supersample\! \subsetchoice} }\lefto[ \frac{1}{n}\!\sum_{i=1}^n\! \ell\lefto(W,Z_i(\bar S_i)\right) \right]}\\
&\geq\! \epsilon 
\end{align}
with probability no larger than $\delta=2\exp(-2\epsilon^2n/(b-a)^2)$ under $P_{W\!\supersample\! \subsetchoice}$. From this it follows that, with probability at least $1-\delta$ under $P_{W\!\supersample\! \subsetchoice}$,
\begin{equation}\label{B_eq:genshat_to_gens}
    \abs{ L_{\trainingdata(\bar\subsetchoice)}(W) - \Exop_{P_Z}\lefto[\ell(W,Z)\right] } \leq \sqrt{\frac{(b-a)^2}{2n}\log\frac{2}{\delta}}.
\end{equation}
Now note that, by the triangle inequality,
\begin{multline}
\abs{\genS} \\
\leq \abs{\genShat} + \abs{ L_{\trainingdata(\bar\subsetchoice)}(W) - \Exop_{P_Z}\lefto[\ell(W,Z)\right] } .
\end{multline}
The result in~\eqref{B_eq:lem_genShat_to_gen_bound} now follows by combining~\eqref{B_eq:pf_of_lem_genShat_condition} and~\eqref{B_eq:genshat_to_gens} via the union bound and performing the substitution $\delta\rightarrow \delta/2$. The proof of~\eqref{B_eq:lem_genShat_to_gen_bound_pb} follows along the same lines.
\end{IEEEproof}

We now turn to proving exponential inequalities similar to Theorem~\ref{thm:unconditional_sub-Gaussian}, but for the CMI setting. These inequalities will later be used to derive generalization bounds of the forms given in~\eqref{B_eq:conditional_exp_gen_error_def},~\eqref{B_eq:conditional_pacb_gen_error_def}, and~\eqref{B_eq:conditional_sd_gen_error_def}.
\begin{thm}\label{thm:conditional_bounded_inequality}
Consider the setting of Theorem~\ref{thm:hoeffding_genshat_to_gens}.
Furthermore, assume that~$P_{W\vert \supersample}$ and~$P_{W\vert \supersample\subsetchoice}$ are absolutely continuous with respect to each other.
Then, for all $\lambda\in \reals$,
\begin{multline}\label{B_eq:conditional_inequality_thm}
    \!\!\!\!\!\Exop_{P_{W\! \supersample\! \subsetchoice}} \lefto[{\exp\lefto(\lambda \genShat\!-\!\frac{\lambda^2(b\!-\!a)^2}{2n}\!-\!\imath(W,\!\subsetchoice\vert \supersample) \! \right)} \right]\\
    \leq 1.
\end{multline}
Furthermore,
\begin{multline}\label{B_eq:conditional_inequality_thm_for_tails}
    \!\!\!\!\!\Exop_{\jointdistro}\biggo[\exp\biggo(\frac{n\!-\!1}{2}(\genShat)^2-\log \sqrt{n} 
    - \condinfdens \!\bigg)\bigg] \\ \leq 1.
\end{multline}
\end{thm}
\begin{IEEEproof}
Due to the boundedness of $\ell(\cdot,\cdot)$, the random variable $\ell(W,Z_i(\bar S_i))-\ell(W,Z_i(S_i))$ is bounded on $[(a-b),(b-a)]$ for $i=1,\dots,n$. As remarked in Lemma~\ref{lem:hoeffdings_inequality}, this implies that it is $(b-a)$-sub-Gaussian, and that $\genShat$ therefore is $(b-a)/\sqrt{n}$-sub-Gaussian. Furthermore, $\genShat$ enjoys the symmetry property $\genShat=-\widehat\genop(W,\supersample, \bar\subsetchoice)$. From this, it follows that $\Exop_{P_{\subsetchoice}}\lefto[\genShat\right]=0$. By the definition of sub-Gaussianity, we therefore have that
\begin{equation}
     \Exop_{P_{\subsetchoice} }\lefto[\exp(\lambda \genShat)\right] \leq \exp\left(\frac{\lambda^2(b-a)^2}{2n} \right).
\end{equation}
Reorganizing terms and taking the expectation with respect to $P_{W\!\supersample}$, we obtain
\begin{equation}\label{B_eq:where_we_can_choose_other_QWZ_in_thm}
\Exop_{P_{W\!\supersample}P_{\subsetchoice} }\lefto[\exp\left(\lambda \genShat - \frac{\lambda^2 (b-a)^2}{2n} \right) \right] \leq 1.
\end{equation}
Due to the absolute continuity assumptions, we can perform a change of measure to $P_{W \! \supersample \! \subsetchoice}$, as per \cite[Prop.~17.1(4)]{PA-polyanskiy19-a}, after which the~\eqref{B_eq:conditional_inequality_thm} follows.

To derive~\eqref{B_eq:conditional_inequality_thm_for_tails}, we note that~\eqref{B_eq:subgauss-square-lemma} with $\lambda=1-1/n$ implies that
\begin{equation}
\Ex{P_{\subsetchoice} }{\exp\lefto(\frac{n-1}{2}(\genShatwz)^2 \right) } \leq \sqrt{n}.
\end{equation}
Taking the expectation with respect to $P_{W\!\supersample}$, changing measure to $\jointdistro$, and rearranging terms, we obtain the desired result.
\end{IEEEproof}
Similar to the discussion in Remark~\ref{rem:uncond_conversion_to_Q_bounds}, Theorem~\ref{thm:conditional_bounded_inequality} holds \emph{verbatim} with $P_{W\vert \supersample}$ replaced by an auxiliary conditional distribution $Q_{W\vert \supersample}$, provided that a suitable absolute continuity assumption holds. This is detailed in the following remark.
While an analogous statement holds for~\eqref{B_eq:conditional_inequality_thm_for_tails}, we present only the version similar to~\eqref{B_eq:conditional_inequality_thm} for brevity.
\begin{remark}\label{rem:cond_conversion_to_Q_bounds}
Consider the setting of Theorem~\ref{thm:hoeffding_genshat_to_gens}. Also, assume that the absolute continuity criteria $P_{W\!\supersample \! \subsetchoice}\ll Q_{W\vert \supersample}P_{\supersample}P_{\subsetchoice}$ and $Q_{W\vert \supersample}P_{\supersample}P_{\subsetchoice} \ll P_{W\!\supersample \! \subsetchoice}$ hold for some conditional distribution $Q_{W\vert \supersample}$ on $\mathcal{W}$. Then,
\begin{multline}\label{B_eq:conditional_inequality_thm_with_Q}
    \Exop_{P_{W\! \supersample\! \subsetchoice}} \biggo[\exp\biggo(\!\lambda \genShat\!-\!\frac{\lambda^2(b\!-\!a)^2}{2n}\!\\
    -\!\log\! \frac{\dv P_{W\! \supersample\! \subsetchoice} }{\dv Q_{W\vert \supersample  }P_{\supersample}P_{\subsetchoice} }\! \bigg) \bigg] 
    \leq 1 .
\end{multline}
The proof is exactly the same as the proof of Theorem~\ref{thm:conditional_bounded_inequality}, except that we use $Q_{W\vert \supersample}$ in place of $P_{W\vert \supersample}$ in the change of measure.
\end{remark}
For the bounds that we will later derive, the optimal choice is $Q_{W\vert \supersample}=P_{W\vert \supersample}$. However, similar to the standard setting, this choice is not always feasible when one is interested in numerically evaluating the bounds. While it is technically possible to compute $P_{W\vert \supersample}$ for a given instance of $\supersample$ by marginalizing out $\subsetchoice$, this would involve executing the probabilistic learning algorithm $P_{W\vert \trainingdata(\subsetchoice)}$ a total of $2^n$ times. For many algorithms, this is prohibitively expensive from a computational standpoint. Therefore, it can be convenient to have the choice of relaxing the bound by expressing it in terms of some auxiliary distribution $Q_{W\vert \supersample}$, suitably chosen so as to trade accuracy with computational complexity.

We also note that the assumption of bounded loss in Theorem~\ref{thm:conditional_bounded_inequality} can be relaxed, and exponential inequalities can be derived for unbounded loss functions satisfying the conditions specified in the following remark.
While an analogous statement holds for~\eqref{B_eq:conditional_inequality_thm_for_tails}, we present only the version similar to~\eqref{B_eq:conditional_inequality_thm} for brevity.
\begin{remark}
Assume that there exists a function $\Delta:\mathcal{Z}^2\rightarrow\reals$ such that, for all $z_1,z_2\in \mathcal{Z}$ and all $w\in\mathcal{W}$, we have $\abs{\ell(w,z_1)-\ell(w,z_2)}\leq \Delta(z_1,z_2)$. Let $Z_1$ and $Z_2$ be independent and distributed according to $P_Z$. Then, for all $\lambda\in \reals$,
\begin{multline}\label{B_eq:conditional_inequality_thm_unbound}
    \Exop_{P_{W\! \supersample\! \subsetchoice}} \biggo[\exp\biggo(\lambda \genShat-\frac{\lambda^2\Ex{P_{Z_1\!Z_2}}{\Delta(Z_1,Z_2)^2}  }{2n}\\
    -\condinfdens  \bigg)\bigg]\leq 1.
\end{multline}
\end{remark}
The proof of~\eqref{B_eq:conditional_inequality_thm_unbound} involves adapting the derivation in \cite[p.~29]{PA-steinke20-a} and using a change of measure argument. All the bounds presented in the remainder of this section for the bounded loss setting admit a counterpart for the unbounded loss setting, obtained by replacing $(b-a)^2$ with $\Ex{P_{Z_1\!Z_2}}{\Delta(Z_1,Z_2)^2}$. Our choice to focus on the case of bounded loss functions in the remainder of this paper is justified by the fact that boundedness of $\Ex{P_{Z_1\!Z_2}}{\Delta(Z_1,Z_2)^2}$ can be proven only for very specific cases (see~\cite[Sec.~5.4--5.6]{PA-steinke20-a}).

In the remainder of this section, we will use Theorem~\ref{thm:conditional_bounded_inequality} to derive an average generalization bound, as well as PAC-Bayesian bounds and single-draw bounds. We start with the average generalization bound.
\subsection{Average Generalization Error Bounds}\label{sec-a:cond_exp}
In the same spirit as Corollary~\ref{cor:exp_uncond}, the following bound on the average generalization error, which is explicit in the CMI $I(W;\subsetchoice\vert \supersample)$, is directly derived from Theorem~\ref{thm:conditional_bounded_inequality}.
\begin{cor}\label{cor:exp_cond}
Under the setting of Theorem~\ref{thm:hoeffding_genshat_to_gens},
\begin{equation}\label{B_eq:cor_exp_cond}
    \abs{ \Ex{P_{W\! \supersample \! \subsetchoice}}{ \textnormal{gen}(W,\trainingdata(\subsetchoice)) }} \leq \sqrt{\frac{2(b-a)^2}{n}I(W;\subsetchoice\vert \supersample)}.
\end{equation}
\end{cor}
\begin{IEEEproof}
Starting from the inequality in~\eqref{B_eq:conditional_inequality_thm}, we apply Jensen's inequality, which results in
\begin{multline}
    \exp\biggo(\lambda\Exop_{P_{W\! \supersample \! \subsetchoice}}\biggo[\genShat\bigg] - \frac{\lambda^2(b-a)^2}{2n}  \\
 -\Ex{P_{W\! \supersample \! \subsetchoice}}{\imath(W,\subsetchoice\vert \supersample)} \bigg)  \leq   1.
\end{multline}
Note that $\Exop_{P_{W\! \supersample \! \subsetchoice}}\lefto[\genShat\right] = \Exop_{P_{W\! \supersample \! \subsetchoice}}\lefto[\textnormal{gen}(W,\trainingdata(\subsetchoice))\right]$, since $W$ and $\trainingdata(\bar\subsetchoice)$ are independent. Also, we have that
\begin{align}
\Exop_{P_{W\! \supersample\! \subsetchoice}}\lefto[\imath(W,\subsetchoice\vert \supersample)\right] &= \conrelent{P_{W\vert \supersample\! \subsetchoice }P_{\subsetchoice} }{P_{W\vert \supersample}P_{\subsetchoice} }{P_{\supersample}}\\
&=I(W;\subsetchoice\vert \supersample).
\end{align}
We therefore get, after reorganizing terms,
\begin{multline}
 \lambda\Exop_{P_{W\! \supersample \! \subsetchoice}}\lefto[\genS\right] - \frac{\lambda^2(b-a)^2}{2n} 
 \\
 -\conrelent{P_{W\vert \supersample\! \subsetchoice }P_{\subsetchoice} }{P_{W\vert \supersample}P_{\subsetchoice} }{P_{\supersample}}  \leq   0.
\end{multline}
Setting $\lambda = n \Exop_{P_{W\! \supersample \! \subsetchoice}}\lefto[\genS\right] / (b-a)^2$ to optimize the bound, we obtain
\begin{multline}\label{B_eq:exp_cond_final_line_of_proof}
    \Exop_{P_{W\! \supersample \! \subsetchoice}}^2\lefto[\genS\right] \\
    - \frac{2(b-a)^2}{n}\conrelent{P_{W\vert \supersample\! \subsetchoice }P_{\subsetchoice} }{P_{W\vert \supersample}P_{\subsetchoice} }{P_{\supersample}} \leq 0
\end{multline}
from which~\eqref{B_eq:cor_exp_cond} follows directly.
\end{IEEEproof}
The bound in~\eqref{B_eq:cor_exp_cond} recovers the result from \cite[Cor.~5.2]{PA-steinke20-a}. As detailed in Remark~\ref{rem:cond_conversion_to_Q_bounds}, we can substitute $Q_{W\vert \supersample}$ for $P_{W\vert \supersample}$ in~\eqref{B_eq:exp_cond_final_line_of_proof} to obtain a more general but weaker bound in terms of the conditional relative entropy $\conrelent{P_{W\vert \supersample\! \subsetchoice }P_{\subsetchoice} }{Q_{W\vert \supersample}P_{\subsetchoice} }{P_{\supersample}}$, provided that an appropriate absolute continuity assumption is satisfied.

Under some conditions, the bound in Corollary~\ref{cor:exp_cond} can be shown to be tighter than that in Corollary~\ref{cor:exp_uncond} for the case of a bounded loss function. Indeed, using the chain rule for mutual information, the Markov property $(\supersample, \subsetchoice)$---$\trainingdata(\subsetchoice)$---$W$, and the fact that $\trainingdata(\subsetchoice)$ is a deterministic function of $(\supersample, \subsetchoice)$, we can rewrite the bound in~\eqref{B_eq:cor_exp_uncond} as
\begin{align}
\abs{\textnormal{gen}(W,\trainingdata(\subsetchoice))} \!&\leq\! \sqrt{ \frac{(b-a)^2}{2n}I(W;
\trainingdata(\subsetchoice))
} \\
&=\! \sqrt{ \frac{(b\!-\!a)^2}{2n}\!\left(\!I(W;\!\supersample)\!+\!I(W;\!\subsetchoice\vert \supersample)\! \right) }.\!
\end{align}
Hence, if $I(W;\supersample)>3I(W;\subsetchoice\vert \supersample)$, the bound in Corollary~\ref{cor:exp_cond} is tighter than that in Corollary~\ref{cor:exp_uncond}. In particular, note that there are many practical scenarios in which the bound in Corollary~\ref{cor:exp_uncond} is vacuous because $I(W;\trainingdata(\subsetchoice))=\infty$. On the contrary, $I(W;\subsetchoice\vert\supersample)\leq n\log 2$.
\subsection{PAC-Bayesian Generalization Error Bounds}\label{sec-a:cond_pacb}
We now turn to PAC-Bayesian bounds of the form given in~\eqref{B_eq:conditional_pacb_gen_error_def}. The next corollary provides bounds that are analogous to those in Corollary~\ref{cor:pacb_uncond}, but for the CMI setting. The bounds in the corollary are novel, and extend known PAC-Bayesian bounds to the CMI setting.
\begin{cor}\label{cor:pacb_cond}
Under the setting of Theorem~\ref{thm:hoeffding_genshat_to_gens}, the following holds with probability at least $1-\delta$ under $P_{\supersample\! \subsetchoice}$:
\begin{multline}\label{B_eq:cor_pacb_cond}
\abs{ \Exop_{P_{W\vert \supersample\! \subsetchoice} } \lefto[\genShat\right] } \\ \leq  \sqrt{\frac{2(b-a)^2}{n-1}\left(\relent{P_{W\vert \supersample\! \subsetchoice}}{P_{W\vert \supersample} } +\log \frac{\sqrt n}{\delta}\right)} %
\end{multline}
\begin{multline}
\label{B_eq:cor_pacb_cond_second}
\abs{ \Exop_{P_{W\vert \supersample\! \subsetchoice} } \lefto[\genShat\right] } \\
\leq\!  \sqrt{\!\frac{2(b\!-\!a)^2}{n}\!\!\left(\frac
    {\Exop^{1/t}_{P_{\supersample\! \subsetchoice} }\lefto[\relent{P_{W\vert \supersample\! \subsetchoice}\!}{\!P_{W\vert \supersample} }^t\right] }
    {(\delta/2)^{1/t}} \!+\!\log \!\frac{2}{\delta}\!\right)}%
\end{multline}
for all $t>0$.
Here, the first inequality is a data-dependent bound, while the second provides a data-independent version.
\end{cor}
\begin{IEEEproof}
Since the proof follows along the same lines as that of Corollary~\ref{cor:pacb_uncond}, we only highlight the differences.
We start from~\eqref{B_eq:conditional_inequality_thm_for_tails}, apply Jensen's inequality with respect to $P_{W\vert \supersample\! \subsetchoice}$, and note that
\begin{equation}
    \Exop_{P_{W\vert \supersample\! \subsetchoice} }\lefto[\condinfdens\right] = \relent{P_{W\vert \supersample\! \subsetchoice}}{P_{W\vert \supersample} }.
\end{equation}
To obtain~\eqref{B_eq:cor_pacb_cond}, we use~\eqref{B_eq:markov_inequality_Pu} and rearrange terms.

Next, to prove~\eqref{B_eq:cor_pacb_cond_second}, we apply Markov's inequality to $\relent{P_{W\vert \supersample\! \subsetchoice}}{P_{W\vert \supersample} }^t$, similarly to~\eqref{B_eq:uncond_pacb_proof_markov_with_t}.
We then combine the resulting inequality with~\eqref{B_eq:conditional_inequality_thm} through the union bound.
Now, we let $a={\Exop^{1/t}_{P_{\supersample\! \subsetchoice} }\lefto[\relent{P_{W\vert \supersample\! \subsetchoice}}{P_{W\vert \supersample} }^t\right]}/{\delta^{1/t}}
+\log \frac{1}{\delta}$ and~$b=2(b-a)^2/n$.
The desired result in~\eqref{B_eq:cor_pacb_cond} follows by setting $\lambda = \pm \sqrt{\frac{a}{b}}$ and performing the substitution $\delta\rightarrow \delta/2$.
\end{IEEEproof}
Now, note that for the case $t=1$ in~\eqref{B_eq:cor_pacb_cond_second}, we have that $\Exop_{P_{\supersample\! \subsetchoice} }\lefto[\relent{P_{W\vert \supersample\! \subsetchoice}}{P_{W\vert \supersample} }\right]=I(W;\subsetchoice\vert \supersample)$. The corresponding bound extends the results in~\cite{PA-steinke20-a} by providing a PAC-Bayesian generalization error bound in terms of the CMI $I(W;\subsetchoice\vert \supersample)$. Similar to the discussion following Corollary~\ref{cor:exp_cond}, this bound is, under some conditions, tighter than the corresponding bounds for the standard setting in Corollary~\ref{cor:pacb_uncond}. Much like the moment bounds in~\eqref{B_eq:cor_pacb_uncond_second} and~\eqref{B_eq:cor_singledraw_uncond_second}, the bound in~\eqref{B_eq:cor_pacb_cond_second} illustrates a trade-off between the confidence and the tightness of the generalization estimate, mediated by the magnitude of the higher moments of $\relent{P_{W\vert \supersample\! \subsetchoice}}{P_{W\vert \supersample} }$.
Again, a data-independent relaxation can be derived directly from~\eqref{B_eq:cor_pacb_cond}, but the resulting bound has a logarithmic dependence on~$n$ that is not present in~\eqref{B_eq:cor_pacb_cond_second}.
Also, as indicated in Remark~\ref{rem:cond_conversion_to_Q_bounds}, if the appropriate absolute continuity criterion is satisfied, we can replace $P_{W\vert\supersample}$ with $Q_{W\vert \supersample}$ in~\eqref{B_eq:cor_pacb_cond} and~\eqref{B_eq:cor_pacb_cond_second} to obtain more general bounds that are better suited for numerical evaluations.
\subsection{Single-Draw Generalization Error Bounds}\label{sec-a:cond_sd}
In this section, we derive several bounds on the single-draw generalization error~\eqref{B_eq:conditional_sd_gen_error_def} in the CMI setting. Three different approaches will be used to obtain these bounds. The first one relies on the exponential inequalities given in Theorem~\ref{thm:conditional_bounded_inequality}, and results in a data-dependent bound along with several data-independent versions. The second one relies on Lemma~\ref{lem:strong_converse_lemma}, and allows us to derive a bound that is explicit in the tail of the conditional information density, similar to Theorem~\ref{thm:unconditional_hypothesis_main_bound}. Essentially equivalent versions of the data-independent bounds obtainable via the first approach can be derived from this tail-based bound. The third approach, which is inspired by~\cite{PA-esposito19-12a}, builds on repeated applications of H\"older's inequality. This results in a family of data-independent bounds. Through this approach, we extend many of the results for bounded loss functions in~\cite{PA-esposito19-12a} to the CMI setting.
\subsubsection{Generalization Bounds from the Exponential Inequalities}
In the next two corollaries, we derive novel bounds that are analogous to the ones in Corollaries~\ref{cor:singledraw_uncond} and~\ref{cor:singledraw_uncond_alphadiv_leakage}, but for the CMI setting.
\begin{cor}\label{cor:singledraw_cond}
Under the setting of Theorem~\ref{thm:hoeffding_genshat_to_gens}, the following holds with probability at least $1-\delta$ under $P_{W\!\supersample\! \subsetchoice}$:\footnote{Note that the argument of the square root in~\eqref{B_eq:cor_singledraw_cond} can be negative, but that this happens with probability at most $\delta$. Therefore, the right-hand side of~\eqref{B_eq:cor_singledraw_cond} is well-defined with probability at least $1-\delta$.}
\begin{equation}\label{B_eq:cor_singledraw_cond}
    \abs{ \genShat } \! \leq \! \sqrt{\!{\frac{2(b\!-\!a)^2}{n\!-\!1}\!\left(\!\imath(W,\subsetchoice\vert \supersample) \!+\!\log \frac{\sqrt n}{\delta}\right)} } 
    \end{equation}
    \begin{multline}\label{B_eq:cor_singledraw_cond_second}
    \abs{ \genShat } \\ \leq \! \sqrt{\!\frac{2(b-a)^2}{n}\!\left(\!I(W;\!\subsetchoice\vert \supersample)  \!+\! \frac{\widetilde M_t(W;\subsetchoice\vert \supersample)}{(\delta/2)^{1/t}} \!+\!\log\! \frac{2}{\delta}\!\right)}\!
\end{multline}
for all $t>0$.
Here, the first inequality provides a data-dependent bound and the second is a data-independent version. In~\eqref{B_eq:cor_singledraw_cond_second}, the term $\widetilde M_t(W;\subsetchoice\vert \supersample)$ is the $t$th root of the $t$th central moment of $\imath(W,\subsetchoice\vert \supersample)$:
\begin{equation}\label{B_eq:def_tilde_M_conditional}
\widetilde M_t(W;\subsetchoice\vert \supersample) \!=\! \Exop^{1/t}_{P_{W\! \supersample \! \subsetchoice }}\lefto[\abs{ \imath(W,\!\subsetchoice\vert \supersample) \!-\! I(W;\!\subsetchoice\vert \supersample) }^t \right].\!
\end{equation}
\end{cor}
\begin{IEEEproof}
The proof is analogous to that of Corollary~\ref{cor:singledraw_uncond}. We start by applying Markov's inequality in the form of~\eqref{B_eq:markov_inequality_Pu} to~\eqref{B_eq:conditional_inequality_thm_for_tails}, which results in~\eqref{B_eq:cor_singledraw_cond} after reorganizing.
For the data-independent version, we first apply~~\eqref{B_eq:markov_inequality_Pu} to~\eqref{B_eq:conditional_inequality_thm}, resulting in
\begin{multline}\label{B_eq:after_markov_cond_sd_base}
\Exop_{P_{W\! \supersample\! \subsetchoice}} \biggo[\exp\biggo(\lambda \genShat-\frac{\lambda^2(b-a)^2}{2n}\\
-\log \frac{\dv P_{W\! \supersample\! \subsetchoice} }{\dv Q_{W\vert \supersample  }P_{\supersample}P_{\subsetchoice} } \bigg) \bigg]\leq 1
\end{multline}
We then apply~\eqref{B_eq:markov_for_arbitrary_u} with $U=\imath(W,\subsetchoice\vert \supersample)$ to~\eqref{B_eq:after_markov_cond_sd_base}.
Now, let $a=\frac{\widetilde M_t(W;\subsetchoice\vert \supersample)}{(\delta)^{1/t}}
+\log \frac{1}{\delta}$ and~$b=2(b-a)^2/n$.
Combining the resulting inequality with~\eqref{B_eq:cor_singledraw_cond} through the union bound, we obtain~\eqref{B_eq:cor_singledraw_cond_second} after setting~$\lambda = \pm \sqrt{\frac{a}{b}}$ and performing the substitution $\delta\rightarrow \delta/2$.
\end{IEEEproof}
By increasing $t$ in~\eqref{B_eq:cor_singledraw_cond_second}, a more benign $\delta$-dependence can be obtained by letting the bound depend on higher central moments of $\condinfdens$. The tightness of the resulting bound depends on how well these higher moments are controlled.
Again, a data-independent relaxation can be derived directly from~\eqref{B_eq:cor_singledraw_cond}, but the resulting bound has a logarithmic dependence on~$n$ that is not present in~\eqref{B_eq:cor_singledraw_cond_second}.
As usual, we can get more general bounds by replacing $P_{W\vert \supersample}$ with an arbitrary $Q_{W\vert \supersample}$, provided that a suitable absolute continuity assumption is satisfied.

Just as in Corollary~\ref{cor:singledraw_uncond_alphadiv_leakage}, we can derive alternative data-independent versions of the data-dependent bound in~\eqref{B_eq:cor_singledraw_cond}. We present these novel bounds in the following corollary. The first bound is given in terms of $\mathcal{L}(\subsetchoice \rightarrow W\vert \supersample)$, the conditional maximal leakage~\eqref{B_eq:def_cond_maximal_leakage}. The second bound depends on the conditional R\'enyi divergence~\eqref{B_eq:def_cond_renyi_divergence}.
\begin{cor}\label{cor:singledraw_cond_alphadiv_leakage}
Under the setting of Theorem~\ref{thm:hoeffding_genshat_to_gens}, the following inequalities hold with probability at least $1-\delta$ under $P_{W\!\supersample\!\subsetchoice}$:
\begin{equation}\label{B_eq:cor_singledraw_cond_leakage}
\abs{ \genShat } \!\leq\! \sqrt{\frac{2(b\!-\!a)^2}{n}\!\!\left(\!\mathcal{L}(\subsetchoice \!\rightarrow \! W\vert \supersample) \!+\!2\!\log\! \frac{2}{\delta}\!\right)}\!
\end{equation}
and, for all $\alpha,\gamma> 1$ such that $1/\alpha+1/\gamma = 1$,
\begin{multline}\label{B_eq:cor_singledraw_cond_alphadiv}
\abs{ \genShat } 
\leq \\
\bigg[\frac{2(b-a)^2}{n}\bigg(\frac{\alpha-1}{\alpha}\alphaconrelent{\alpha}{P_{W\vert \supersample\!\subsetchoice}P_{\subsetchoice} }{P_{W\vert \supersample}P_{\subsetchoice}}{P_{\supersample}} \\+ \frac{\gamma\!-\!1}{\gamma}\alphaconrelent{\gamma}{P_{W\vert \supersample\!\subsetchoice}P_{\subsetchoice} }{P_{W\vert \supersample}P_{\subsetchoice}}{P_{\supersample}}\!+\!2\log\! \frac{2}{\delta}\bigg)\!\bigg]^{1/2}\!\!.\!
\end{multline}
\end{cor}
\begin{IEEEproof}
Analogously to the proof of Corollary~\ref{cor:singledraw_uncond_alphadiv_leakage}, we start from the inequality in~\eqref{B_eq:after_markov_cond_sd_base} and bound $\imath(W,\subsetchoice\vert \supersample)$. Markov's inequality implies that, with probability $1-\delta$~under $P_{W\!\supersample\! \subsetchoice}$,
\begin{equation}\label{B_eq:singledraw_cond_leakage_esssup_weakening}
\imath(W,\subsetchoice\vert \supersample) = \log \Ex{P_{W\!\supersample\!\subsetchoice} }{\frac{\dv P_{W\! \supersample\! \subsetchoice} }{\dv P_{W\vert \supersample  }P_{\supersample\! \subsetchoice} }} +\log \frac{1}{\delta}.
\end{equation}
Replacing expectations with essential suprema, we get the upper bound
\begin{align}
&\Ex{P_{W\!\supersample\!\subsetchoice} }{\frac{\dv P_{W\! \supersample\! \subsetchoice} }{\dv P_{W\vert \supersample  }P_{\supersample\! \subsetchoice} }} \nonumber\\
& \leq   \esssup_{P_{\supersample} }\Ex{P_{W\vert \supersample}}{ \esssup_{P_{\subsetchoice\vert W\!\supersample}} \frac{\dv P_{W\! \supersample\! \subsetchoice} }{\dv P_{W\vert \supersample  }P_{\supersample\! \subsetchoice} } } \\\label{B_eq:cor_cond_sd_leakage_proof_infdens_bound_last}
& \leq   \esssup_{P_{\supersample} }\Ex{P_{W\vert \supersample}}{ \esssup_{P_{\subsetchoice}} \frac{\dv P_{W\! \supersample\! \subsetchoice} }{\dv P_{W\vert \supersample  }P_{\supersample\! \subsetchoice} } }.
\end{align}
Here, the second inequality holds due to the absolute continuity property $P_{W\!\supersample \! \subsetchoice}\ll P_{W\vert \supersample}P_{\supersample\! \subsetchoice}$.
Using the union bound to combine the probabilistic inequality on $\condinfdens$ resulting from~\eqref{B_eq:singledraw_cond_leakage_esssup_weakening}--\eqref{B_eq:cor_cond_sd_leakage_proof_infdens_bound_last} with~\eqref{B_eq:after_markov_cond_sd_base}, %
the result in~\eqref{B_eq:cor_singledraw_cond_leakage} follows after analogous steps to the proof of~\eqref{B_eq:cor_singledraw_cond_second}.

To prove~\eqref{B_eq:cor_singledraw_cond_alphadiv}, we apply Markov's inequality and then perform a change of measure from $P_{W\vert \supersample\!\subsetchoice}$ to $P_{W\vert \supersample}$ to conclude that, with probability at least $1-\delta$ under $P_{W\! \supersample\! \subsetchoice}$,
\begin{align}
\imath(W,\subsetchoice\vert \supersample) & \leq  \log \Exop_{P_{W\vert \supersample \! \subsetchoice}P_ {\supersample\! \subsetchoice}}\lefto[ \frac{\dv P_{W\! \supersample\! \subsetchoice} }{\dv P_{W\vert \supersample  }P_{\supersample\! \subsetchoice} } \right] + \log\frac{1}{\delta}\\
&= \log\! \Exop_{P_{W\vert \supersample }P_ {\supersample\! \subsetchoice}}\!\lefto[\!\left( \!\frac{\dv P_{W\! \supersample\! \subsetchoice} }{\dv P_{W\vert \supersample  }P_{\supersample\! \subsetchoice} } \!\right)^{\!2}\right] \!+\! \log\!\frac{1}{\delta}.\!\label{B_eq:infodens_bound_by_holder}
\end{align}
Next, we apply H\"older's inequality thrice as follows. Let $\alpha,\gamma,\alpha',\gamma',\tilde \alpha, \tilde \gamma> 1$ be constants such that $1/\alpha+1/\gamma=1/\alpha'+1/\gamma'=1/\tilde \alpha + 1/\tilde \gamma=1$. Then,%
\begin{align}
& \Exop_{P_{W\vert \supersample }P_ {\supersample\! \subsetchoice}}\!\lefto[\!\left(\!\! \frac{\dv P_{W\! \supersample\! \subsetchoice} }{\dv P_{W\vert \supersample  }P_{\supersample\! \subsetchoice} } \!\right)^{\!\!2}\right] 
\!\!=\! \Exop_{P_{W\vert \supersample }P_ {\supersample}P_{\subsetchoice}}\lefto[\!e^{2\condinfdens}\!\right]\!\! \\
& \leq  \Exop_{P_{W\vert \supersample }P_ {\supersample}}\biggo[\Exop_{P_{\subsetchoice}}^{1/\alpha}
\lefto[e^{\alpha\condinfdens} \right]\cdot \Exop_{P_{\subsetchoice}}^{1/\gamma}
\lefto[e^{\gamma\condinfdens} \right]
\bigg]  \\
& \leq  \Exop_{P_ {\supersample}}\biggo[\Exop_{P_{W\vert \supersample } }^{1/\tilde \alpha}\lefto[ \Exop_{P_{\subsetchoice}}^{\tilde \alpha/\alpha}
\lefto[e^{\alpha\condinfdens} \right]\right]\\
&\qquad\qquad\quad\nonumber \cdot \Exop_{P_{W\vert \supersample } }^{1/\tilde \gamma}\lefto[\Exop_{P_{\subsetchoice}}^{\tilde \gamma/\gamma}
\lefto[e^{\gamma\condinfdens} \right] \right]
\bigg]  \\  \label{B_eq:holdering_gives_alphadiv}
& \leq\! \Exop_{P_ {\supersample}}^{1/\alpha'}\!\biggo[\!\Exop_{P_{W\vert \supersample } }^{\alpha'/\tilde \alpha}\!\lefto[ \Exop_{P_{\subsetchoice}}^{\tilde \alpha/\alpha}
\!\lefto[e^{\alpha\condinfdens} \right]\!\right]\!\!\bigg]\\ 
&\qquad\qquad\quad\nonumber \cdot \Exop_{P_ {\supersample}}^{1/\gamma'}\!\biggo[\! \Exop_{P_{W\vert \supersample } }^{\gamma'/\tilde \gamma}\!\lefto[\Exop_{P_{\subsetchoice}}^{\tilde \gamma/\gamma}
\lefto[e^{\gamma\condinfdens} \right]\! \right]\!\!
\bigg]\!.\! \!
\end{align}
We now substitute~\eqref{B_eq:holdering_gives_alphadiv} into~\eqref{B_eq:infodens_bound_by_holder} and set $\alpha=\alpha'=\tilde \alpha$, which implies $\gamma=\gamma'=\tilde \gamma$. Using~\eqref{B_eq:def_cond_renyi_divergence}, we conclude that, with probability at least $1-\delta$ under $P_{W\!\supersample\!\subsetchoice}$,
\begin{multline}\label{B_eq:infodens_bounded_by_alpha_divs}
\imath(W,\subsetchoice\vert \supersample)   \leq \frac{\alpha-1}{\alpha}\alphaconrelent{\alpha}{P_{W\vert \supersample\! \subsetchoice}P_{\subsetchoice} }{P_{W\vert \supersample}P_{\subsetchoice}}{P_{\supersample}} \\+ \frac{\gamma-1}{\gamma}\alphaconrelent{\gamma}{P_{W\vert \supersample\! \subsetchoice}P_{\subsetchoice} }{P_{W\vert \supersample}P_{\subsetchoice}}{P_{\supersample}}+\log\frac{1}{\delta}.
\end{multline}
Combining~\eqref{B_eq:infodens_bounded_by_alpha_divs} with~\eqref{B_eq:after_markov_cond_sd_base} through the union bound, the result in~\eqref{B_eq:cor_singledraw_cond_alphadiv} follows after analogous steps to the proof of~\eqref{B_eq:cor_singledraw_cond_second}.
\end{IEEEproof}
As usual, we can replace $P_{W\vert \supersample}$ by some auxiliary $Q_{W\vert \supersample}$ to get more general bounds, provided that a suitable absolute continuity assumption is satisfied.
Furthermore, data-independent relaxations in terms of the conditional maximal leakage and the conditional R\'enyi divergence can be obtained directly from~\eqref{B_eq:cor_singledraw_cond_second}, but the resulting bound has a logarithmic dependence on~$n$ that is not present in Corollary~\ref{cor:singledraw_cond_alphadiv_leakage}.

The conditional maximal leakage bound in~\eqref{B_eq:cor_singledraw_cond_leakage} can be tighter than the maximal leakage bound in~\cite[Cor.~9]{PA-esposito19-12a}.\footnote{Note that $\eqref{B_eq:cor_singledraw_cond_leakage}$ provides a bound on $\genShat$, whereas the bound in~\cite[Cor.~9]{PA-esposito19-12a} is on $\gen$. To compare the two, one therefore has to add the $\delta$-dependent penalty term in Theorem~\ref{thm:hoeffding_genshat_to_gens}.} This is the case since the conditional maximal leakage $\mathcal{L}(\subsetchoice\rightarrow W \vert \supersample)$ is upper-bounded by the maximal leakage $\mathcal{L}(\trainingdata(\subsetchoice)\rightarrow W )$. We prove this result in the following theorem.
\begin{thm}\label{propo:leakage_bounds_cond_leakage}
Consider the setting of Theorem~\ref{thm:hoeffding_genshat_to_gens}. Then,
\begin{equation}\label{B_eq:propo_leakage_bounds_cond_leakage}
\mathcal{L}(\subsetchoice\rightarrow W \vert \supersample) \leq \mathcal{L}( \trainingdata(\subsetchoice) \rightarrow W).
\end{equation}
\end{thm}
\begin{IEEEproof}
Because of the Markov property $(\supersample, \subsetchoice)$---$\trainingdata(\subsetchoice)$---$W$ and the fact that $\trainingdata(\subsetchoice)$ is a deterministic function of $(\supersample,\subsetchoice)$, the equality $\mathcal{L}(\trainingdata(\subsetchoice) \rightarrow W) =\mathcal{L}((\supersample,\subsetchoice) \rightarrow W)$ holds~\cite[Lem.~1]{PA-issa16-a}. We begin by moving one essential supremum outside of the expectation:
\begin{align}\label{B_eq:cond_max_leak_prop_proof_eq_1}
\mathcal{L}((\supersample,\!\subsetchoice)\!\rightarrow\! W) \!&=\! \log \Exop_{P_W}\lefto[ \esssup_{P_{\supersample\! \subsetchoice}} \frac{\dv P_{W\! \supersample\!\subsetchoice}}{\dv P_W \! P_{\supersample\! \subsetchoice} }\right]
\\
&\!\geq\! \log \esssup_{P_{\supersample} } \Exop_{P_W}\!\lefto[ \esssup_{P_{\subsetchoice}} \frac{\dv P_{W\! \supersample\!\subsetchoice}}{\dv P_W \! P_{\supersample\! \subsetchoice} }\right]\!.\!
\end{align}
Now, let $\setE_{\supersample}=\textnormal{supp}(P_{W\vert \supersample})$. It follows from~\eqref{B_eq:cond_max_leak_prop_proof_eq_1} that
\begin{multline}
\mathcal{L}((\supersample,\subsetchoice)\rightarrow W) \\
\geq \log \esssup_{P_{\supersample} } \Exop_{P_W}\lefto[ 1_{\setE_{\supersample} } \esssup_{P_{\subsetchoice}} \frac{\dv P_{W\! \supersample\! \subsetchoice}}{\dv P_W \! P_{\supersample\! \subsetchoice} }\right].
\end{multline}
Next, we perform a change of measure from $P_W$ to $P_{W\vert\supersample}$:
\begin{multline}
 \esssup_{P_{\supersample} } \Exop_{P_W}\lefto[ 1_{\setE_{\supersample} } \esssup_{P_{\subsetchoice}} \frac{\dv P_{W\! \supersample\!\subsetchoice}}{\dv P_W \! P_{\supersample\! \subsetchoice} }\right] \\
 =  \esssup_{P_{\supersample} } \Exop_{P_{W\vert \supersample} }\lefto[\frac{\dv P_W}{\dv P_{W\vert \supersample} } \esssup_{P_{\subsetchoice}} \frac{\dv P_{W\! \supersample\!\subsetchoice}}{\dv P_W \! P_{\supersample\! \subsetchoice} }\right]\!.
\end{multline}
Finally, since $\dv P_W/\dv P_{W\vert \supersample}$ is independent of $\subsetchoice$,
\begin{align}\nonumber
&\log \esssup_{P_{\supersample} } \Exop_{P_{W\vert \supersample} }\lefto[\frac{\dv P_W}{\dv P_{W\vert \supersample} } \esssup_{P_{\subsetchoice}} \frac{\dv P_{W\! \supersample\!\subsetchoice}}{\dv P_W \! P_{\supersample\! \subsetchoice} }\right]\\
&=\log \esssup_{P_{\supersample} } \Exop_{P_{W\vert \supersample} }\lefto[ \esssup_{P_{\subsetchoice}} \frac{\dv P_{W\! \supersample\!\subsetchoice}}{\dv P_{W\vert \supersample}P_{\supersample\! \subsetchoice} }\right]\\
&=\mathcal{L}(\subsetchoice\rightarrow W\vert \supersample).
\end{align}
\end{IEEEproof}
\subsubsection{Generalization Bounds from the Strong Converse}\label{sec-a:cond_sd_strong_conv}
In this section, we will use Lemma~\ref{lem:strong_converse_lemma} to derive single-draw generalization error bounds in the CMI setting. In Theorem~\ref{thm:conditional_hypothesis_main_bound} below, we use Lemma~\ref{lem:strong_converse_lemma} to obtain a novel bound in terms of the tail of the conditional information density $\condinfdens$.
\begin{thm}\label{thm:conditional_hypothesis_main_bound}
Under the setting of Theorem~\ref{thm:hoeffding_genshat_to_gens}, with probability at least $1-\delta$ under $P_{W\! \supersample\! \subsetchoice}$,
\begin{multline}\label{B_eq:thm_conditional_hypothesis_main_bound}
    \abs{\genShat}
    \leq\\\sqrt{ \!\frac{2(b\!-\!a)^2}{n}\!
    \!\left(\!
        \gamma \!+\! \log\lefto(\!\frac{2}{\delta\!-\!P_{W\! \supersample \! \subsetchoice}\lefto[\imath(W,\subsetchoice\vert \supersample)\!\geq\! \gamma\right]} \right)\!
    \right)}.\!
\end{multline}
This is valid for all $\gamma$ such that the right-hand side is defined and real.
\end{thm}
\begin{IEEEproof}
We will use Lemma~\ref{lem:strong_converse_lemma} with $P\!=\!P_{W\! \supersample \! \subsetchoice}$, $Q\!=\!P_{W\vert \supersample}P_{\supersample\! \subsetchoice}$ and
\begin{equation}\label{B_eq:high_error_event_cond}
\setE=\{W,\supersample, \subsetchoice: \abs{\genShat}>\epsilon \}.
\end{equation}
Let the set $\setE_{W\! \supersample}=\{\subsetchoice: (W,\supersample,\subsetchoice)\in \setE \}$ denote the fibers of $\setE$ with respect to $W$ and $\supersample$. As noted in the proof of Theorem~\ref{thm:conditional_bounded_inequality}, $\genShat$ is a $(b-a)/\sqrt{n}$-sub-Gaussian random variable with $\Exop_{P_{\subsetchoice}}\lefto[\genShat \right]=0$. By using Lemma~\ref{lem:hoeffdings_inequality}, we therefore conclude that, for all $W$ and $\supersample$,
\begin{equation}\label{B_eq:mcdiarmid_es_bound}
P_{\subsetchoice}\lefto[\setE_{W\!\supersample} \right] \leq 2 \exp\lefto(-\frac{n\epsilon^2}{2(b-a)^2} \right).
\end{equation}
From this, it follows that $Q[\setE]\leq  2\exp\lefto(-n\epsilon^2/2(b-a)^2 \right)$. By inserting this inequality into~\eqref{B_eq:lem_strong_converse_lemma}, we get
\begin{multline}\label{B_eq:strong_conv_cond_solve_for_eps}
    P_{W\! \supersample \! \subsetchoice}\lefto[\abs{\genShat}>\epsilon\right] \\
    \leq\! P_{W\! \supersample \! \subsetchoice}\lefto[\imath(W,\subsetchoice\vert \supersample)\!\geq\! \gamma\right] \!+\! 2\exp\lefto(\gamma\!-\!\frac{n\epsilon^2}{2(b\!-\!a)^2}\right).
\end{multline}
We obtain the desired result by requiring the right-hand side of~\eqref{B_eq:strong_conv_cond_solve_for_eps} to equal $\delta$ and solving for $\epsilon$.%
\end{IEEEproof}
Similar to the discussion in Remark~\ref{rem:cond_conversion_to_Q_bounds}, a completely analogous result holds with an auxiliary distribution $Q_{W\vert \supersample}$ in place of $P_{W\vert \supersample}$, provided that a suitable absolute continuity assumption is satisfied.

As for the bound in Theorem~\ref{thm:unconditional_hypothesis_main_bound}, the bound in~\eqref{B_eq:thm_conditional_hypothesis_main_bound} illustrates that the faster the rate of decay of the tail of the conditional information density, the sharper the generalization bound. Specifically, the parameter $\gamma$ has to be chosen large enough so that the argument of the logarithm is positive, but a greater $\gamma$ also contributes to an increased value for the bound.

The bound in Theorem~\ref{thm:conditional_hypothesis_main_bound} can be relaxed to give essentially equivalent versions of some of the previously presented data-independent bounds. We show this in the following remarks.
\begin{remark}[Alternative derivation of the moment bound~\eqref{B_eq:cor_singledraw_cond_second}]
Markov's inequality implies that
\begin{align}\label{B_eq:moment_bound_from_hyptest_markov_cond}
P_{W\! \supersample\! \subsetchoice}\lefto[\imath(W,\subsetchoice\vert \supersample) \geq \gamma\right]& \leq  \frac{(\widetilde M_t(W;\subsetchoice\vert \supersample))^t }{\left(\gamma-I(W;\subsetchoice\vert \supersample)\right)^{t}}
\end{align}
where $\widetilde M_t(W;\subsetchoice\vert \supersample)$ is defined in~\eqref{B_eq:def_tilde_M_conditional}. Next, we set
\begin{equation}%
    \gamma =I(W;\subsetchoice\vert \supersample)+\frac{\widetilde M_t(W;\subsetchoice\vert \supersample)}{(\delta/2)^{1/t}}
\end{equation}
which, once it is substituted into~\eqref{B_eq:moment_bound_from_hyptest_markov_cond}, implies that we have $P_{W\! \supersample\! \subsetchoice}\lefto[\imath(W,\subsetchoice\vert \supersample)\geq \gamma \right] \leq {\delta}/{2}$. Using this inequality in~\eqref{B_eq:thm_conditional_hypothesis_main_bound}, we conclude that, with probability at least $1-\delta$ under $P_{W\!\supersample\!\subsetchoice}$,
\begin{multline}
    \abs{\genShat}
    \leq\\
    \sqrt{ \!\frac{2(b\!-\!a)^2}{n}\!
    \left(\!
        I(W;\!\subsetchoice\vert \supersample)\!+\!\frac{\widetilde M_t(W;\!\subsetchoice\vert \supersample)}{(\delta/2)^{1/t}} \!+\! \log\!\frac{4}{\delta}\!
    \right)}.\!
\end{multline}
This coincides with~\eqref{B_eq:cor_singledraw_cond_second}, up to a $(2(b-a)^2/n)\log 2$ term inside the square root.
\end{remark}
\begin{remark}[Alternative derivation of the conditional maximal leakage bound~\eqref{B_eq:cor_singledraw_cond_leakage}]
Note that
\begin{align}
&P_{W\! \supersample \! \subsetchoice}\lefto[\imath(W,\subsetchoice\vert \supersample)\geq \gamma\right] \nonumber\\
& \leq\!  P_{W\!\supersample}\lefto[\esssup_{P_{\subsetchoice\vert W\!\supersample}}
\exp\lefto(\condinfdens \right)>e^\gamma\right]\\
& \leq\!  \esssup_{P_{\supersample}}P_{W\vert\supersample}\!\lefto[\esssup_{P_{\subsetchoice\vert W\!\supersample}}
\exp\lefto(\condinfdens \right)\!>\!e^\gamma\right].
\end{align}
By upper-bounding the $\esssup$ as in~\eqref{B_eq:cor_cond_sd_leakage_proof_infdens_bound_last} and using Markov's inequality, we conclude that
\begin{align}
&P_{W\! \supersample \! \subsetchoice}\lefto[\imath(W,\subsetchoice\vert \supersample)\geq \gamma\right]  \nonumber\\
& \leq   e^{-\gamma} \esssup_{P_{\supersample} } \Ex{P_{W\vert \supersample}}{\esssup_{P_{\subsetchoice}}\exp\lefto(\condinfdens \right) 
}\\
&= \exp\lefto(\mathcal{L}(\subsetchoice \rightarrow W\vert \supersample)-\gamma\right).
\end{align}
Setting $\gamma=\mathcal{L}(\subsetchoice \rightarrow W\vert \supersample) + \log(2/\delta)$ and substituting the resulting upper-bound on the probability $P_{W\! \supersample \! \subsetchoice}\lefto[\imath(W,\subsetchoice\vert \supersample)\geq \gamma\right]$ into~\eqref{B_eq:thm_conditional_hypothesis_main_bound}, we conclude that, with probability at least $1-\delta$ under $P_{W\!\supersample\!\subsetchoice}$,
\begin{multline}
\abs{ \genShat } \leq \\
\sqrt{\frac{2(b-a)^2}{n}\left(\mathcal{L}(\subsetchoice \rightarrow W\vert \supersample)+\log 2 + 2\log \frac{2}{\delta}\right)}.
\end{multline}
This recovers the conditional maximal leakage bound in~\eqref{B_eq:cor_singledraw_cond_leakage}, up to a term $(2(b-a)^2/n)\log 2$ inside the square root.
\end{remark}
\subsubsection{Generalization Bounds from a H\"older-Based Inequality}
We now present a third approach to obtain data-independent single-draw bounds in the CMI setting. The approach is based on a proof technique developed in~\cite{PA-esposito19-12a}, where similar bounds are derived in the standard setting. We first prove a useful inequality in Theorem~\ref{thm:esposito_cond}, from which several generalization bounds follow.
\begin{thm}\label{thm:esposito_cond}
Under the setting of Theorem~\ref{thm:hoeffding_genshat_to_gens}, for all $\alpha, \gamma,\alpha',\gamma',\tilde \alpha, \tilde \gamma> 1$ such that $1/\alpha + 1/\gamma =1/\alpha'+1/\gamma'=1/\tilde \alpha + 1/\tilde \gamma =1$ and all measurable sets $\setE\in \mathcal{W}\times \mathcal{Z}^{2n}\times \{0,1\}^n$,
\begin{multline}\label{B_eq:thm_esposito_cond}
P_{W\! \supersample\! \subsetchoice}[\setE] 
\!\leq\! \Exop_{P_{\supersample} }^{1/\tilde \gamma}\lefto[ \Exop_{P_{W\vert \supersample}}^{\tilde \gamma/\gamma '}\!\lefto[P_{\subsetchoice}^{\gamma'/\gamma}\lefto[\setE_{W\! \supersample}\right] \right] \right] \\ \cdot  \Exop_{P_{\supersample}}^{1/\tilde \alpha}\lefto[ \Exop_{P_{W\vert \supersample}}^{\tilde \alpha/\alpha '}\lefto[\Exop_{P_S}^{\alpha ' /\alpha}\!\lefto[\exp\lefto(\alpha\imath(W,\subsetchoice\vert\supersample)\right) \right] \!\right] \right]\!.\!
\end{multline}
Here, $\setE_{W\!\supersample}=\{\subsetchoice: (W,\supersample,\subsetchoice)\in \setE \}$ denotes the fibers of $\setE$ with respect to $W$ and $\supersample$.
\end{thm}
\begin{IEEEproof}
First, we rewrite $P_{W\! \supersample\! \subsetchoice}[\setE] $ in terms of the expectation of the indicator function $1_\setE$ and perform a change of measure:
\begin{align}
P_{W\!\supersample\! \subsetchoice}[\setE]\!&=\! \Exop_{P_{W\vert \supersample }P_{\supersample\! \subsetchoice}} \lefto[1_\setE \cdot \frac{\dv P_{W\! \supersample\! \subsetchoice} }{\dv P_{W\vert \supersample} P_{\supersample\! \subsetchoice} } \right]\! \\
&=\! \Exop_{ P_{W\vert \supersample}  P_{\supersample } P_{\subsetchoice}} \lefto[ 1_{\setE} \cdot exp\lefto(\imath(W,\subsetchoice\vert\supersample)\right) \right].
\end{align}
To obtain the desired result, we apply H\"older's inequality thrice. Let $\alpha, \gamma,\alpha',\gamma',\tilde \alpha, \tilde \gamma> 1$ be constants such that $1/\alpha + 1/\gamma =1/\alpha'+1/\gamma'=1/\tilde \alpha + 1/\tilde \gamma =1$. Then,
\begin{align}
{P_{W\! \supersample\! \subsetchoice}}[\setE] 
\!& \leq\!\Exop_{P_{W\vert \supersample}P_{\supersample}}\lefto[ \Exop_{P_{\subsetchoice}}^{1/\gamma}\lefto[1_{\setE_{W\! \supersample}}\right]\cdot\Exop_{P_{\subsetchoice}}^{1/\alpha}\lefto[ e^{\alpha\imath(W,\subsetchoice\vert\supersample)} \right] \right]\nonumber\\
\!& \leq\!\Exop_{P_{\supersample}} \biggo[ \Exop_{P_{W\vert \supersample}}^{1/\gamma'}\lefto[P_{\subsetchoice}^{\gamma'/\gamma}\lefto[\setE_{W\! \supersample}\right] \right]
\\&\qquad\qquad\cdot \Exop_{P_{W\vert \supersample}}^{1/\alpha'}\lefto[\Exop_{P_{\subsetchoice}}^{\alpha'/\alpha}\lefto[ e^{\alpha\imath(W,\subsetchoice\vert\supersample)} \right] \right]  \bigg] \nonumber\\ 
\!& \leq\!\Exop_{P_{\supersample}}^{1/\tilde \gamma}\lefto[ \Exop_{P_{W\vert \supersample}}^{\tilde \gamma/\gamma'}\!\lefto[P_{\subsetchoice}^{\gamma'/\gamma}\lefto[\setE_{W\! \supersample}\right] \right] \right]  \\
&\qquad\qquad\cdot \Exop_{P_{\supersample}}^{1/\tilde \alpha}\!\lefto[ \Exop_{P_{W\vert {\supersample}}}^{\tilde \alpha/\alpha'}\!\lefto[\Exop_{P_{\subsetchoice}}^{\alpha'/\alpha}\lefto[e^{\alpha\imath(W,\subsetchoice\vert\supersample)} \right] \right] \right]\!.\!\nonumber
\end{align}
\end{IEEEproof}
Similar to the discussion in Remark~\ref{rem:cond_conversion_to_Q_bounds}, the result in Theorem~\ref{thm:esposito_cond} would still hold if we were to substitute an auxiliary distribution $Q_{W\vert \supersample}$ for $P_{W\vert \supersample}$, provided that a suitable absolute continuity condition is satisfied. 

By choosing particular values for the three free parameters in the inequality~\eqref{B_eq:thm_esposito_cond}, we can derive generalization bounds in terms of various information-theoretic quantities. We will focus on a bound that depends on the conditional $\alpha$-mutual information $\alphaconMI{\alpha}{W }{\subsetchoice}{\supersample }$, which can be relaxed to obtain a bound in terms of the conditional R\'enyi divergence $\alphaconrelent{\alpha}{P_{W\vert \supersample\!\subsetchoice}P_{\subsetchoice} }{P_{W\vert \supersample}P_{\subsetchoice}}{P_{\supersample}}$ or be specialized to obtain a bound that depends on the conditional maximal leakage $\mathcal{L}(\subsetchoice \rightarrow W\vert \supersample)$.
\begin{cor}\label{cor:singledraw_esposito_cond_alphaMI}
Under the setting of Theorem~\ref{thm:hoeffding_genshat_to_gens}, the following holds with probability at least $1-\delta$ under $P_{W\! \supersample\! \subsetchoice}$ for all $\alpha > 1$:
\begin{multline}\label{B_eq:cor_singledraw_esposito_cond_alphaMI}
\abs{ \genShat } \leq\\
\sqrt{\frac{2(b-a)^2}{n}\left(\alphaconMI{\alpha}{W }{\subsetchoice}{\supersample }+\log 2 + \frac{\alpha}{\alpha-1}\log \frac{1}{\delta}\right)}.
\end{multline}
\end{cor}
\begin{IEEEproof}
In~\eqref{B_eq:thm_esposito_cond}, set $\tilde \alpha = \alpha$ and let $\alpha' \rightarrow 1$, which implies that $\tilde \gamma = \gamma$ and $\gamma' \rightarrow\infty$. Also, let $\setE$ be the error event~\eqref{B_eq:high_error_event_cond}. For this choice of parameters, the second factor in~\eqref{B_eq:thm_esposito_cond} reduces to
\begin{multline}\label{B_eq:esposito_alphami_proof_info_dens_factor}
\Exop_{P_{\supersample}}^{1/\alpha}\lefto[ \Exop^{\alpha}_{P_{W\vert \supersample}}\lefto[\Exop^{1/\alpha}_{P_S}\lefto[\exp\lefto(\alpha\imath(W,\subsetchoice\vert\supersample)\right) \right] \right] \right] \\
= \exp\lefto(\frac{\alpha-1}{\alpha}\alphaconMI{\alpha}{W }{\subsetchoice}{\supersample }\right).
\end{multline}
Furthermore, we can bound $P_{\subsetchoice}\lefto[\setE_{W\! \supersample}\right]$ in the first factor in~\eqref{B_eq:thm_esposito_cond} by using~\eqref{B_eq:mcdiarmid_es_bound}. Substituting~\eqref{B_eq:mcdiarmid_es_bound} into the first factor in~\eqref{B_eq:thm_esposito_cond}, we conclude that
\begin{align}
&\lim_{\gamma' \rightarrow\infty} \Exop_{P_{\supersample}}^{1/ \gamma}\lefto[ \Exop_{P_{W\vert \supersample}}^{ \gamma/\gamma'}\lefto[P^{\gamma'/\gamma}_{\subsetchoice}\lefto[\setE_{W\! \supersample}\right] \right] \right]  \nonumber\\
&= \Exop^{1/\gamma}_{P_{\supersample}}\lefto[ \lefto(\esssup_{P_{W\vert \supersample}}P^{1/\gamma}_{\subsetchoice}\lefto[\setE_{W\! \supersample}\right]\right)^{\gamma} \right]  \\
& \leq  \left(  2 \exp\lefto(-\frac{n\epsilon^2}{2(b-a)^2} \right)\right)^{1/\gamma}.\label{B_eq:esposito_alphami_proof_E_dep_factor}
\end{align}
By substituting~\eqref{B_eq:esposito_alphami_proof_info_dens_factor} and~\eqref{B_eq:esposito_alphami_proof_E_dep_factor} into~\eqref{B_eq:thm_esposito_cond}, noting that $1/\gamma = (\alpha-1)/\alpha$, we conclude that
\begin{multline}\label{B_eq:deriv_cond_alpha_mi_last}
P_{W\! \supersample \! \subsetchoice}[\setE]
\leq \left( 2 \exp\lefto(-\frac{n\epsilon^2}{2(b-a)^2} \right) \right)^{\frac{\alpha-1}{\alpha} }  \\
\cdot  \exp\lefto(\frac{\alpha-1}{\alpha}\alphaconMI{\alpha}{W }{\subsetchoice}{\supersample }\right).
\end{multline}
We obtain the desired result by requiring the right-hand side of~\eqref{B_eq:deriv_cond_alpha_mi_last} to equal $\delta$ and solving for $\epsilon$.
\end{IEEEproof}
As usual, we can obtain a more general version of Corollary~\ref{cor:singledraw_esposito_cond_alphaMI} by replacing $P_{W\vert \supersample}$ with an auxiliary distribution $Q_{W\vert \supersample}$, provided that a suitable absolute continuity assumption is satisfied.

We can also obtain a bound in terms of the conditional maximal leakage by letting $\alpha\rightarrow \infty$ in~\eqref{B_eq:cor_singledraw_esposito_cond_alphaMI} and using that $\lim_{\alpha\rightarrow\infty} \alphaconMI{\alpha}{W }{\subsetchoice}{\supersample }= \mathcal{L}(\subsetchoice \rightarrow W\vert \supersample)$. The resulting bound is tighter than the conditional maximal leakage bound obtained in~\eqref{B_eq:cor_singledraw_cond_leakage} by a $(2(b-a)^2/n)\log (2/\delta)$ term inside the square root.

Furthermore, the conditional $\alpha$-mutual information that appears in~\eqref{B_eq:cor_singledraw_esposito_cond_alphaMI} can be relaxed to obtain a novel bound in terms of the conditional R\'enyi divergence of order $\alpha$. Indeed, by Jensen's inequality, the following holds for $\alpha>1$:
\begin{align}
&\alphaconMI{\alpha}{W}{\subsetchoice}{\supersample} \nonumber\\\label{B_eq:cond_alpha_mi_to_cond_renyi_start}
&=\frac{1}{\alpha-1}\log \Exop_{P_ {\supersample}}\lefto[\Exop_{P_{W\vert \supersample } }^{\alpha}\lefto[ \Exop_{P_{\subsetchoice}}^{1/\alpha}
\lefto[e^{\alpha\condinfdens } \right]\right]\right] \\
& \leq  \frac{1}{\alpha-1}\log \Exop_{P_ {\supersample}}\lefto[\Exop_{P_{W\vert \supersample } }\lefto[ \Exop_{P_{\subsetchoice}}
\lefto[e^{\alpha\condinfdens } \right]\right]\right]\\
&= \alphaconrelent{\alpha}{P_{W\vert \supersample\!\subsetchoice}P_{\subsetchoice} }{P_{W\vert \supersample}P_{\subsetchoice}}{P_{\supersample}}.\label{B_eq:cond_alpha_mi_to_cond_renyi_end}
\end{align}
The conditional R\'enyi divergence bound obtained by substituting~\eqref{B_eq:cond_alpha_mi_to_cond_renyi_end} into~\eqref{B_eq:cor_singledraw_esposito_cond_alphaMI} is different from the one in~\eqref{B_eq:cor_singledraw_cond_alphadiv}, and there is no clear ordering between them in general. The two bounds can, however, be directly compared if we set $\alpha=\gamma=2$, or if we let $\alpha\rightarrow \infty$, and hence $\gamma \rightarrow 1$. For both of these choices of parameters, the conditional R\'enyi divergence bound obtained from~\eqref{B_eq:cor_singledraw_esposito_cond_alphaMI} is tighter than~\eqref{B_eq:cor_singledraw_cond_alphadiv} by a $(2(b-a)^2/n)\log (2/\delta)$ term inside the square root.

\section{Conclusion}
We have presented a general framework for deriving generalization bounds for probabilistic learning algorithms, not only in the average sense, but also for the PAC-Bayesian and the single-draw setup. Using this framework, we recovered several known results, and also presented new ones.
While the concept of exponential inequalities has been used before, for instance in~\cite{PA-catoni07-a} and~\cite{PA-koolen-16a}, we clarified the different flavors of bounds that this leads to and extended this perspective to the CMI setting and sub-Gaussian losses for the standard setting.
Due to its unifying nature, the framework enables the transfer of methods for tightening bounds in one setup to the other two setups. %
In particular, by reobtaining previously known results, we showed that our framework subsumes proofs that are based on the Donsker-Varadhan variational formula for relative entropy~\cite[Thm.~1]{PA-xu17-05a},~\cite[Prop.~3]{PA-guedj19-10a}, on H\"older's inequality~\cite[Thm.~1]{PA-esposito19-12a}, and on the data-processing inequality~\cite[Thm.~8]{PA-bassily18-02a},~\cite[p.~10]{PA-esposito19-12a}. We further demonstrated the versatility of the framework by applying it to the CMI setting recently introduced by Steinke and Zakynthinou~\cite{PA-steinke20-a}. In doing so, we were able to extend the bounds on the average generalization error obtained in~\cite{PA-steinke20-a} to the PAC-Bayesian setup and the single-draw setup.
We note that, while we used two different exponential inequalities to derive our results for each setting, it is possible to derive results of all three flavors from a single exponential inequality.
However, some of the resulting bounds would have a logarithmic dependence on~$n$ that can be avoided through the use of two inequalities, as pointed out repeatedly throughout this paper.
In addition to this, we used tools inspired by binary hypothesis testing to derive generalization bounds in terms of the tail of the conditional information density. We also obtained novel bounds in terms of the conditional maximal leakage and the conditional $\alpha$-mutual information by adapting a proof technique due to Esposito \textit{et al.}~\cite{PA-esposito19-12a} to the CMI setting.

As pointed out throughout, the numerical evaluation of the presented generalization bounds often requires one to replace the marginal distribution $P_W$ (or $P_{W\vert\supersample}$ in the CMI setting) with a suitably chosen auxiliary distribution that can be computed without \textit{a priori} knowledge of the data distribution $P_Z$. Some possible choices, in the context of deep neural networks, are provided in \cite{PA-Negrea2019,PA-Dziugaite2017,PA-dziugaite-20,PA-hellstrom-20-c}. Specifically, in \cite{PA-hellstrom-20-c}, we evaluate the PAC-Bayesian bound in~\eqref{B_eq:cor_pacb_cond} and the single-draw bound in~\eqref{B_eq:cor_singledraw_cond} for neural networks trained on MNIST and Fashion-MNIST using stochastic gradient descent. The numerical experiments illustrate that the resulting bounds are non-vacuous for the setups considered, and match the best bounds available in the literature~\cite{PA-dziugaite-20}. While the results in \cite{PA-hellstrom-20-c} appear promising, they still do not provide much insight into how to design neural networks. Thus, the extent to which information-theoretic bounds such as the ones presented in this paper can guide the design of modern machine learning algorithms remains to be investigated.

\section*{Acknowledgements}
The authors would like to thank Pradeep Banerjee, Mahdi Haghifam, Amedeo Roberto Esposito, Max Raginsky, and Yury Polyanskiy who all helped to identify errors in an earlier version of this paper.

\bibliographystyle{IEEEtran}
\bibliography{ReferencesPA}

\end{document}